\documentclass[lettersize,journal]{IEEEtran}
\usepackage{amsmath,amsfonts}
\usepackage{algorithmic}
 \usepackage[ruled, vlined, linesnumbered]{algorithm2e}
\usepackage{array}
\usepackage[caption=false,font=footnotesize,labelfont=rm,textfont=rm]{subfig}
\usepackage{textcomp}
\usepackage{stfloats}
\usepackage{url}
\usepackage{verbatim}
\usepackage{graphicx}
\usepackage{cite}
\usepackage{pifont} 
\usepackage{tabularx}
\usepackage{makecell}
\usepackage{multirow}
\usepackage{booktabs}
\usepackage[colorlinks,citecolor=blue, linkcolor=red, urlcolor=black]{hyperref}
\begin{document}
\title{Exploiting Inherent Class Label: Towards Robust Scribble Supervised Semantic Segmentation }
\author{Xinliang Zhang, 
Lei Zhu, 
Shuang Zeng, 
Hangzhou He, 
Ourui Fu, 
Zhengjian Yao, 
Zhaoheng Xie,
Yanye Lu*
\thanks{This paper was supported in part by the Natural Science Foundation of China under Grant 62394311, 82371112, 623B2001,62394314.

Xinliang Zhang, Lei Zhu, Shuang Zeng, Hangzhou He,  Ourui Fu, Zhengjian Yao, Zhaoheng Xie and Yanye Lu are with Institute of Medical Technology, Peking University Health Science Center, Peking University, Beijing, and also with Department of Biomedical Engineering, Peking University, Beijing, China, and also with National Biomedical Imaging Center, Peking University, Beijing 100871, China, and also with Peking University Shenzhen Graduate School, Shenzhen 518055, China. Yanye Lu is the corresponding author, email:yanye.lu@pku.edu.cn
}
}

\markboth{Journal of \LaTeX\ Class Files,~Vol.~14, No.~8, August~2021}%
{Shell \MakeLowercase{\textit{et al.}}: A Sample Article Using IEEEtran.cls for IEEE Journals}


\maketitle

\begin{abstract}
Scribble-based weakly supervised semantic segmentation leverages only a few annotated pixels as labels to train a segmentation model, presenting significant potential for reducing the human labor involved in the annotation process. This approach faces two primary challenges: first, the sparsity of scribble annotations can lead to inconsistent predictions due to limited supervision; second, the variability in scribble annotations, reflecting differing human annotator preferences, can prevent the model from consistently capturing the discriminative regions of objects, potentially leading to unstable predictions. To address these issues, we propose a holistic framework, the class-driven scribble promotion network, for robust scribble-supervised semantic segmentation. This framework not only utilizes the provided scribble annotations but also leverages their associated class labels to generate reliable pseudo-labels. Within the network, we introduce a localization rectification module to mitigate noisy labels and a distance perception module to identify reliable regions surrounding scribble annotations and pseudo-labels. In addition, we introduce new large-scale benchmarks, ScribbleCOCO and ScribbleCityscapes, accompanied by a scribble simulation algorithm that enables evaluation across varying scribble styles. Our method demonstrates competitive performance in both accuracy and robustness, underscoring its superiority over existing approaches. The datasets and the codes will be made publicly available.
\end{abstract}


\section{Introduction}
Semantic segmentation is a fundamental problem in computer vision, which has seen rapid advancements in recent years, largely driven by the availability of large-scale, pixel-level annotated datasets~\cite{pascal_voc,lin2014microsoft}. However, the creation of such datasets through manual annotation remains labor-intensive and time-consuming, limiting the broader applicability of semantic segmentation techniques. To address these challenges, weakly supervised semantic segmentation methods leveraging sparse labels have gained prominence. These methods utilize annotations at varying levels, such as image level~\cite{zhu2023branches,zhu2023background,ru2023token}, scribble level~\cite{xu2021scribble,pan2021scribble,wu2023sparsely,liang2022tree,Scribble023Zhang}, or bounding box level~\cite{dai2015boxsup,papandreou2015weakly,khoreva2017simple,zhang2021affinity}. Among these, image-level annotations provide minimal supervision, while bounding boxes often suffer from overlaps when objects are in close proximity. In contrast, scribble annotations offer an optimal balance between training supervision and annotation effort~\cite{lin2016scribblesup}, making Scribble-supervised Semantic Segmentation (SSSS) an increasingly popular research focus~\cite{liang2022tree,wu2023sparsely}. 
\begin{figure}
    \centering
    \includegraphics[width=\linewidth]{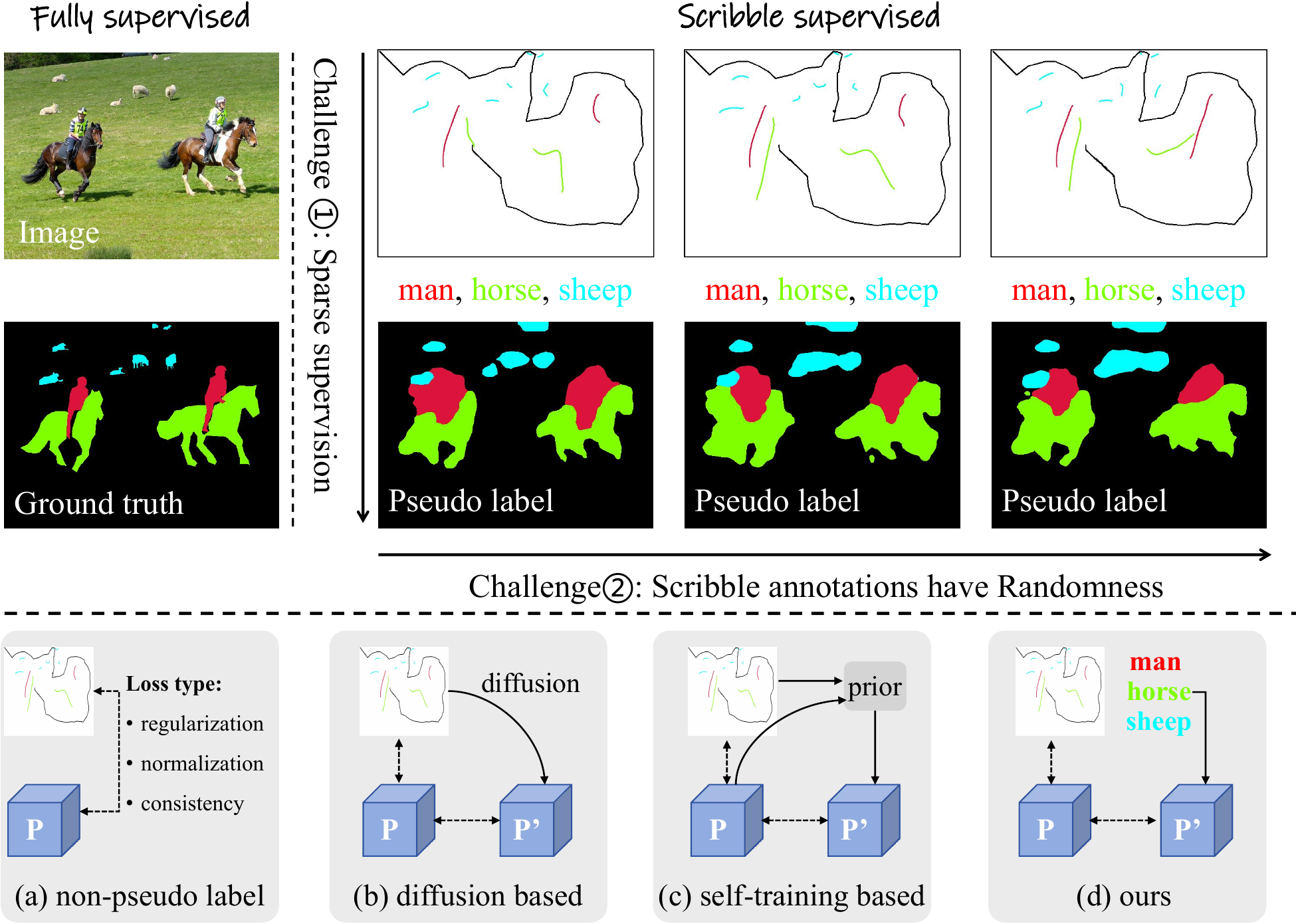}
    \caption{Scribble-supervised semantic segmentation presents two main challenges. While the issue of sparse supervision can be addressed by introducing pseudo-labels, the robustness challenge posed by the variability of scribble annotations remains to be investigated. In order to mitigate this issue, we proposed to generate the pseudo label via the inherent class label in scribble as presented in (d), which is different from previous methods (a)-(c). $P$ represents the model's prediction, $P'$ represents the pseudo label, the dotted line represents the supervision. }
    \label{fig:two_problem}
\end{figure}

The primary challenge in SSSS arises from the sparsity of labels, which provides insufficient supervision and leads to inconsistent predictions. Classical approaches attempt to mitigate this issue by introducing regularization losses~\cite{tang2018normalized,tang2018regularized} to improve generalization. However, these methods do not fully address the supervision deficiency, resulting in suboptimal segmentation performance. More recently, pseudo-label based methods have demonstrated significant progress in improving segmentation accuracy by providing denser supervision. Depending on their pseudo-label generation strategies, these methods can be broadly categorized into two groups: iteratively propagate sparse scribble labels into dense labels~\cite{xu2021scribble,wang2022cycle,zhou2023exploratory}, or enhancing model predictions as pseudo-labels for self-supervision~\cite{liang2022tree,wu2023sparsely}. The former approach relies on local-to-local propagation, which is limited by the original shape of the scribble and often fails to capture entire object regions. The latter approach, on the other hand, derives pseudo-labels from model predictions, making the model prone to the Matthew Effect~\cite{chen2022debiased}, where well-performing categories improve further, while poorly performing ones degrade. Another critical challenge in SSSS is the inherent randomness of scribble annotations, as illustrated in Fig.~\ref{fig:two_problem}. In practice, annotations are provided by different annotators, whose habits and preferences may vary. Consequently, in addition to segmentation accuracy, stability is a key metric for evaluating SSSS methods. However, this perspective remains underexplored due to the lack of datasets containing diverse styles of scribble annotations.

To address these challenges, we propose a novel approach to enhance SSSS by leveraging the inherent class labels associated with scribble annotations. Different from the previous works~\cite{xu2021scribble,wang2022cycle,zhou2023exploratory,liang2022tree,wu2023sparsely}, we neither employ label diffusion nor rely on model predictions for pseudo-label generation. Instead, we introduce an additional network to generate pseudo-labels, which is trained using both pixel-level supervision from scribbles and image-level supervision from class labels. In detail, we adopt the image level weakly supervised method~\cite{ru2023token} as the basic network to deal with the image-level supervision, and we further add a shallow FOV~\cite{chen2017deeplab} on the feature embeddings to involve the scribble as pixel-level supervision. Such an operation has three advantages: 1) \textbf{Robust pseudo-label generation}: The image-level supervision operates independently of the scribble, providing a stable basis for generating pseudo-labels and naturally mitigating the impact of scribble shape randomness. 2) \textbf{Improved pseudo-label accuracy}: Incorporating pixel-level supervision from scribble annotations enhances the precision of the generated pseudo-labels. 3) \textbf{Prevention of the Matthew Effect}: By employing an additional network for pseudo-label generation, rather than self-training, the model is less prone to the Matthew Effect, ensuring more balanced performance across all categories.

Simply adopting pseudo-labels can negatively impact the model's accuracy due to the presence of noisy labels. We empirically observe that, noisy labels predominantly occur in the foreground when different objects are in close proximity. To address this issue, we propose a localization rectification module (LoRM), which leverages the similarity of feature representations among foreground pixels belonging to the same semantic class to correct features distorted by noisy labels. Additionally, we introduce a distance perception module (DPM), which performs entropy reduction based on pixel distances, excavating the reliable regions surrounding both the scribble annotations and pseudo-labels. 

An overview of our method is presented in Fig.~\ref{fig:cdsp}, where we refer to it as a Class-driven Scribble Promotion network (CSPNet). In the first stage, we train the scribble-promoted ToCo with both image-level class supervision and pixel-level scribble supervision. Once training is complete, we extract class activation maps (CAMs) from its classification layer and generate pseudo-labels by following the standard pipeline for image-level segmentation methods~\cite{ru2023token}. In the second stage, a semantic segmentation network is trained using both scribble and pseudo labels, with supervision further enhanced by our proposed LoRM and DPM. In a nutshell, our contributions can be summarized as the following:

\begin{itemize}
  \item For the SSSS problem, we propose a class-driven scribble promotion network. In this network, both the image-level class information and the scribble are exploited to generate pseudo-labels that take global factors into account.
  \item  To tackle the issue of noisy labels present in pseudo-labels, we further design a localization rectification module as well as a distance perception module.
  \item  Extensive experiments are carried out on multiple datasets. The results show that our method can achieve competitive segmentation performance across different scribble styles, highlighting the superiority of our proposed approach in terms of both accuracy and robustness.
  \item In this work, we also introduce two novel large-scale benchmarks for the SSSS field, named ScribbleCOCO (approximately 8 times the scale of ScribbleSup) and ScribbleCityscapes. Additionally, we put forward our random scribble simulation algorithm.
\end{itemize}

This paper extends our previous conference version~\cite{Scribble023Zhang} with several significant improvements. First, we enhance the image-level weakly supervised model by incorporating pixel-level supervision from scribble annotations, generating more accurate pseudo-labels and improved overall performance. Second, we introduce two new benchmarks for scribble-supervised semantic segmentation, along with a scribble simulation algorithm that can be applied to any existing mask-level annotated dataset. These newly proposed benchmarks feature multiple scribble styles, providing a more comprehensive evaluation of the robustness and accuracy of SSSS methods.

\section{Related Work}

\subsection{Scribble-supervised semantic segmentation}
In early eras, scribble-based segmentation was performed in an interactive manner~\cite{rother2004grabcut,grady2006random}, where multiple continuous interactions were required to extract object masks, and manually assigned the mask with the corresponding category. As the deep learning technique burst into bloom, scribble-based segmentation methods nowadays adopt the end-to-end training strategy, automatically performing the semantic segmentation. The first work in this field is proposed by Lin et al. named ScribbleSup~\cite{lin2016scribblesup}. This method generated pseudo-labels by optimizing the GrabCut algorithm\cite{rother2004grabcut} to propagate scribble annotations to unlabeled pixels and subsequently used the pseudo-labels to train a DeepLab-MSc-CRF-LargeFOV model~\cite{liang2015semantic}. Later, Tang et al.~\cite{tang2018normalized,tang2018regularized} proposed to address the SSSS problem from the perspective of semi-supervised learning, where the regularization loss~\cite{tang2018regularized} and the normalization loss~\cite{tang2018normalized} were proposed to assist the partial cross-entropy loss from the scribble. More recently, URSS~\cite{pan2021scribble} proposed to address the SSSS problem from the perspective of the model's certainty and consistency. They proposed a network embedded with a random walk module and a superpixel-based minimum entropy loss for uncertainty reduction, as well as a neural eigenspace self-supervision approach to promote consistency. Although these additional loss functions enhanced model stability, the lack of supervision for unlabeled pixels remained a critical limitation, resulting in constrained segmentation accuracy.

Recently, pseudo-label based approaches proposed to address this problem via devising various pseudo-label generation strategies, which can be divided into label-diffusion based~\cite{vernaza2017learning,xu2021scribble,wang2022cycle,zhou2023exploratory} and self-training based approaches~\cite{liang2022tree,wu2023sparsely}. RAWKS~\cite{vernaza2017learning} defined the label-diffusion process in terms of random-walk hitting probabilities and adopted the edge information and the scribble as the propagation boundary conditions to generate the pseudo labels. PSI~\cite{xu2021scribble} proposed a contextual pattern propagation and semantic label diffusion module to iteratively propagate the labeled pixels of the scribble to unlabeled pixels. CCL~\cite{wang2022cycle} proposed to make predictions for unlabeled regions under the guidance of the scribble with the cycle-consistency learning strategy. EIL~\cite{zhou2023exploratory} introduced the exploratory inference learning strategy to iteratively extend the scribble to dense pseudo-labels. Nevertheless, the label-diffusion process is a local-to-local process, which relies heavily on the initial scribble seed. Self-training based methods follow the paradigm of semi-supervised learning, where the model's prediction is adopted as the pseudo label for the next iterative training step. TEL~\cite{liang2022tree} introduced the tree energy loss to assist the model to predict more accurate pseudo-labels. AGMM~\cite{wu2023sparsely} involved the Gaussian prior to adaptively refining the model's prediction with the scribble. Despite such a learning strategy is convenient to implement, the inherent risk of falling into the Matthew Effect~\cite{chen2022debiased} still exists, and overfitting the noisy label generated by the model itself will prevent the model from being better optimized. Specifically, BGP~\cite{wang2019boundary} proposed to involve edge detection as an auxiliary task for complementary supervision of the scribble, while extra boundary labeling is needed.

In conclusion, existing pseudo-label based approaches either heavily depend on the initial shape of the scribble or suffer from overfitting to noisy labels. Furthermore, while most methods primarily focus on generating high-quality pseudo-labels, few address the critical challenge of effectively utilizing these pseudo-labels, especially given the unavoidable presence of noise in them.

\subsection{Weakly supervised semantic segmentation methods}
Besides the scribble, image-level class~\cite{kolesnikov2016seed,wei2017object,zhang2021complementary,zhu2022bagging,zhu2022dawsol}, point-level~\cite{bearman2016s,chen2021seminar,wu2022deep,liang2022tree,wu2023sparsely}, and bounding box-level~\cite{dai2015boxsup,papandreou2015weakly,khoreva2017simple,zhang2021affinity} are also common annotations in weakly supervised semantice segmentation. Among these methods, we treat the point annotation as a subset of the scribble, which will be discussed in the experiments. Bounding boxes, while providing more supervision, also include too much noisy supervision especially when different objects are close to each other. 

As mentioned before, the scribble shape has randomness, which can influence the pseudo-label generation. In contrast, the image-level class label inherent in the scribble is shape-nonrelevant, which motivates us to draw inspiration from image-level weakly supervised semantic segmentation methods. The success of early deep learning-based methods in image classification~\cite{simonyan2014very} spurred numerous efforts in feature visualization. The CAMs~\cite{zhou2016learning} was first introduced to visualize discriminative localization by applying global average pooling to deep features. This technique inspired various approaches to generate pseudo semantic labels from CAM for training segmentation networks~\cite{kolesnikov2016seed,wei2017object,zhang2021complementary,zhu2022bagging,zhu2022dawsol}. SEAM~\cite{wang2020self} proposed a pixel correlation module to refine predictions by considering similar neighboring pixels. AFA~\cite{ru2022learning} addressed image-level supervised semantic segmentation with transformers~\cite{dosovitskiy2020image}, leveraging the long-range modeling capabilities of multi-head self-attention. Recently, Zhu et al.~\cite{zhu2023branches} proposed a branch mutual promotion network to perform the classification and segmentation task in one network. ToCo~\cite{ru2023token} is an accurate transformer-based architecture for weakly supervised semantic segmentation, which developed a patch token contrast module and a class token contrast module to capture both low-level and high-level semantics. It performs the pseudo-label generation and the segmentation in a one-stop manner, which is convenient for training and inference. Overall, the inherent ability of these methods to capture global information makes image-level supervised semantic segmentation a promising avenue to enhance scribble-supervised semantic segmentation.

\subsection{Scribble simulation algorithms}
Recently, scribble has aroused awareness in various fields, showcasing its flexibility and convenience in various application scenarios. Valvano et al.~\cite{valvano2021learning} first introduced the scribble annotation into the medical semantic segmentation field. They provided the scribble annotations on multiple medical semantic segmentation datasets, including ACDC~\cite{olivier2018acdc}, LVSC~\cite{suinesiaputra2014collaborative}, and CHAOS~\cite{suinesiaputra2014collaborative}. Among them, ACDC was manually annotated with experienced expertise. For LVSC and CHAOS, they simulated the scribbles by adopting a random walk algorithm and skeleton extraction algorithm on the ground truth mask respectively. However, the simulation results were coarse, which has a huge gap with the human-annotated ones. ScribblePrompt~\cite{wong2025scribbleprompt} proposed to simulate the scribble by breaking the objects' skeletons or boundaries into lines for interactive segmentation. Scribble4All~\cite{boettcher2024scribbles} proposed to simulate the scribble by connecting the central points along the medial axis of the objects. 

In conclusion, existing simulation methods primarily focus on line-style scribbles. However, in practice, human annotators often prefer to label around object boundaries, resulting in circular-style annotations. Moreover, these methods overlook an important randomness factor: different annotators may exhibit labeling biases for the same object. Consequently, the simulated scribbles generated by these methods remain fixed, failing to capture the variability inherent in real-world annotations and limiting the ability to fairly evaluate the robustness of different models.
 \begin{figure*}
    \centering
    \includegraphics[width=0.99\linewidth]{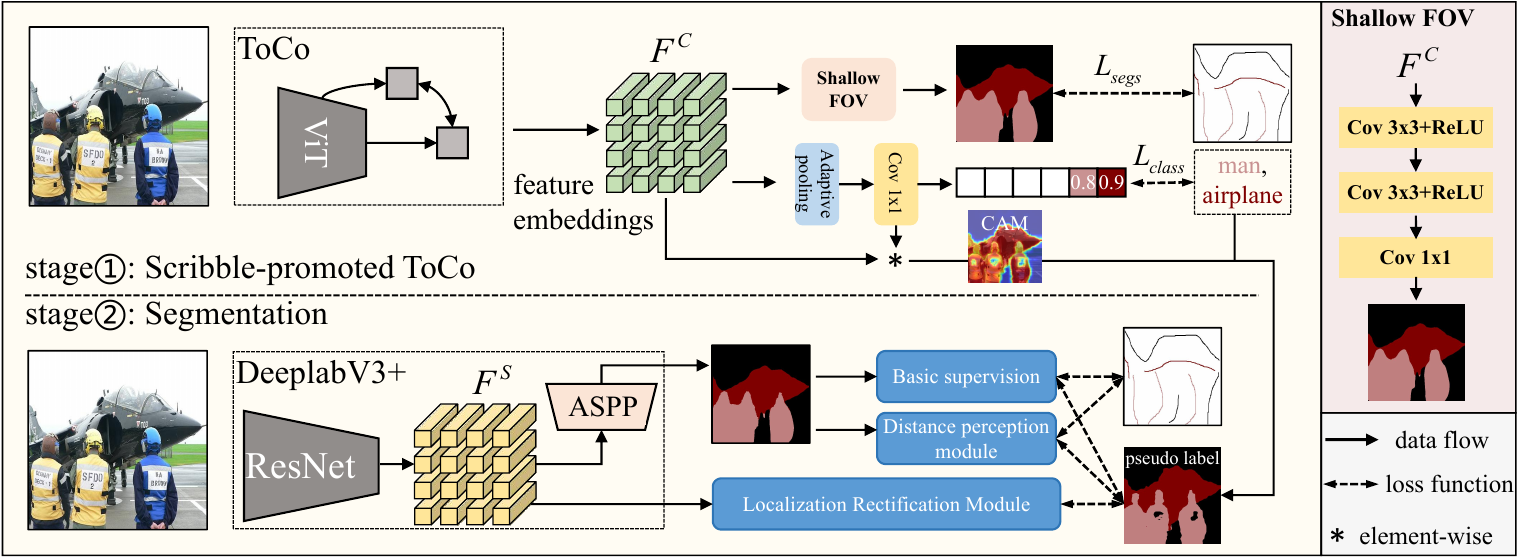}
    \caption{An overview of our method (CSPNet). In the first stage, taking ToCo as the basic network, we train the model with both image-level class supervision and the pixel-level scribble supervision. After training, we generate the pseudo-label from the classification branch, which will be used to train the semantic segmentation model in the second stage. In the second stage, we train the DeeplabV3+ with the basic supervision, distance perception module, and localization rectification module, adopting both scribble and the pseudo-label as the supervision.
}
    \label{fig:cdsp}
\end{figure*}

\section{Class-driven scribble promotion network}
In this section, we first review the general definition of the problem of pseudo-label based methods and discuss their limitations. Following this, we introduce our proposed CSPNet, detailing its components: pseudo-label generation, the localization rectification module, and the distance perception module. An overview of our method is presented in Fig.~\ref{fig:cdsp}. The codes will be made publicly available\footnote{https://github.com/Zxl19990529/CSP-Net, code:``ayri''}.

\subsection{Problem retrospect and basic supervision}
\label{sec:method}
Denoting $\mathbf{\Omega}=\{\boldsymbol{y}_i|i=1,...n\}$ as the ground truth label set and $\mathbf{\Omega}_{s}$ as the sparse scribble label, where $\mathbf{\Omega}_{s} \subset \mathbf{\Omega}$ and $|\mathbf{\Omega}_{s}| << |\mathbf{\Omega}|$.  The objective function of the scribble-supervised semantic segmentation can be formulated as:
\begin{equation}
  \min  c(\mathbf{P}_{\mathbf{\Omega}_s},\mathbf{\Omega}_s),
  \label{eq:ssss}
\end{equation}
where $c(\cdot, \cdot)$ denotes the criterion function, which is usually formulated as a partial cross-entropy. $\mathbf{P}_{\mathbf{\Omega}_s}$ denotes the model predictions corresponding to the sparse scribble label. Training with such sparse supervision will lead to uncertain predictions. Current pseudo-label based methods adopt the strategy of scribble inference to generate the pseudo-labels, which can be concluded as:
\begin{equation}
  \widetilde{\mathbf{\Omega}} = \phi(\mathbf{\Omega}_{s}, \delta),
  \label{eq:diffuse}
\end{equation}
where $\widetilde{\mathbf{\Omega}}=\{\boldsymbol{\widetilde{y}}_i|i=1,...n\}$ is the dense pseudo label, $\phi(\cdot,\delta)$ represents the inference strategy, such as random walk~\cite{vernaza2017learning}, adaptive gaussian predictor~\cite{wu2023sparsely}, pattern propagation module~\cite{xu2021scribble}. For label-diffusion based methods, $\delta$ can be a specific prior, for instance, the edge information~\cite{vernaza2017learning,zhou2023exploratory}, the mean and variance of the confidence map extracted from the model's feature map~\cite{xu2021scribble}, etc. For self-training based methods, the prior $\delta$ can be directly the model's prediction~\cite{liang2022tree}, or the refined prediction with the Gaussian prior involved~\cite{wu2023sparsely}.
Combined with Eq.~\ref{eq:ssss}, a complete objective function for scribble-based WSSS can be obtained:

\begin{equation}
  \min (c(\mathbf{P}_{\mathbf{\Omega}_s},\mathbf{\Omega}_s) +c(\mathbf{P}_{\mathbf{\widetilde{\Omega}}},\mathbf{\widetilde{\Omega}})).
  \label{eq:fsss}
\end{equation}

From Eq.~\ref{eq:diffuse}, the quality of the pseudo-label is unstable, which is affected by two variable factors, the scribble $\mathbf{\Omega}_{s}$, and the prior $\delta$. The scribble has randomness which makes it hard to build a universal prior to generating the pseudo-label accurately, and adopting the model's prediction as the prior may lead to the Matthew Effect~\cite{chen2022debiased}.

To address the problems mentioned above, we propose to generate the pseudo-label from constant factors, which can be formulated as:

\begin{equation}
    \boldsymbol{\widetilde{\Omega}} = \phi(\mathbf{I},\boldsymbol{k}),
    \label{eq:pseudo}
\end{equation}
where $\mathbf{I}$ denotes the input image, $\boldsymbol{k}$ denotes the  image-level class label. Both $\mathbf{I}$ and $\boldsymbol{k}$ are constant, which are non-relevant to the scribble shape, as well as the predictions of the model. As for $\phi$, we adopt the same inference strategy as ToCo~\cite{ru2023token}, which will be introduced in detail in the next section.
After that, we further introduce the LoRM and DPM to strike the advantages of both supervisions as shown in Fig.~\ref{fig:cdsp}. In general, the overall loss function for supervision can be formulated as:
\begin{equation}
    \mathcal{L} = \mathcal{L}_{basic} + \mathcal{L}_{lorm} + \mathcal{L}_{dp}.
    \label{eq:overall_loss}
\end{equation}
In Eq.~\ref{eq:overall_loss}, $\mathcal{L}_{basic}$ represents the basic supervision from the sparse scribble and the dense pseudo-label:
\begin{equation}
  \mathcal{L}_{basic} = \mathcal{L}_{segs} + \lambda_{segc}*\mathcal{L}_{segc},
  \label{eq:l_seg}
\end{equation}
where $\mathcal{L}_{segs}$ is the sparse supervision from the scribble label, $\lambda_{segc}$ is the hyper-parameters to control the weight of the corresponding loss. The formulation of $\mathcal{L}_{segs}$ is a partial cross-entropy:
\begin{equation}
    \mathcal{L}_{segs} = \frac{1}{|\mathbf{\Omega}_s|} \sum_{\boldsymbol{y}_i\in\mathbf{\Omega}_s} c(\boldsymbol{y}_i,\boldsymbol{p}_i),
    \label{eq:l_segs}
\end{equation}
where $c(\boldsymbol{y}_i,\boldsymbol{p}_i)=-\sum_{k=1}^{K} \boldsymbol{y}_{i,k} log(\boldsymbol{p}_{i,k})$, $K$ is the class number, $\boldsymbol{p}_i$ is the prediction from the model, $\boldsymbol{y}_i$ is the one-hot label. $\mathcal{L}_{segc}$ denotes the supervision from the pseudo-label generated by the class, which can be formulated as a regularized cross-entropy:
\begin{equation}
    \mathcal{L}_{segc} = \frac{1}{|\mathbf{\widetilde{\Omega}}|} \sum_{\boldsymbol{y}_i\in \mathbf{\widetilde{\Omega}}}[(1-\epsilon)c(\boldsymbol{y}_i,\boldsymbol{p}_i)+\epsilon c(\frac{1}{K},\boldsymbol{p}_i)],
\end{equation}
where $\epsilon=0.2$ is a regularization item of label smoothing~\cite{muller2019does} to prevent the model from overfitting the pseudo label. 

In Eq.~\ref{eq:overall_loss}, $\mathcal{L}_{lorm}$ represents the supervision from the LoRM, and $\mathcal{L}_{dp}$ is the supervision from DPM. The LoRM is designed to rectify mispredicted pixels by employing a weighted aggregation of representations from neighboring pixels, thereby mitigating the risk of overfitting to noisy labels. The DPM focuses on identifying reliable predictions by leveraging both the scribble and pseudo-label boundaries, thereby enhancing the model's prediction confidence. The details of LoRM and DPM will be introduced sequentially in the following parts.

\subsection{Scribble-promoted ToCo and pseudo label generation}
To acquire the $\phi$ in Eq.~\ref{eq:pseudo}, we presented the scribble-promoted ToCo, as presented in Fig.~\ref{fig:cdsp} in stage 1. The original ToCo is a classification model with the ViT~\cite{dosovitskiy2020image} backbone, where the classification head is a k-channel convolutional kernel with a kernel size of 1. Given the k-class label, the model is trained with soft margin loss, where we denote it as $\mathcal{L}_{cls}$. To further improve the fit of ToCo for SSSS problem, we also involve the pixel-level scribble supervision for training. In detail, we added a shallow version of FOV~\cite{liang2015semantic} as the segmentation head to the original ToCo. Denoting $\mathcal{L}_{segs}$ as the sparse supervision from the scribble label, which is the same calculation with Eq.~\ref{eq:l_segs}, the total loss function of the scribble-promoted ToCo can be formulated as:
\begin{equation}
    \mathcal{L}_{ToCo}=\lambda_{cls}*\mathcal{L}_{cls}+\lambda_{segs} *\mathcal{L}_{segs},
    \label{eq:l_toco}
\end{equation}
where $\lambda_{cls}$ and $\lambda_{segs}$ are the hyper-parameters. It has to be noted that, other auxiliary loss terms are kept as the original ToCo, which we omitted here to keep it concise.

After training, the image $\mathbf{I}$ is fed into the model to generate the class activate map of the $k^{th}$ class:
\begin{equation}
    CAM_{k} (\mathbf{I}) = ReLU( \sum_{i=1}^{C} \mathbf{W}_{i,k}\mathbf{F}_{i}),
\end{equation}
where $\mathbf{F^C}$ is the feature embeddings of the backbone, $\mathbf{W}$ is the weight matrix in the classifier. We follow common image-supervised semantic segmentation methods to threshold the $CAM$ into binary masks and integrate them into a single channel multi-class mask~\cite{chen2022class,ru2023token}. Following the original ToCo, the Pixel-Adaptive Refinement module~\cite{ru2022learning} is adopted to further generate the pseudo label.

\begin{figure}
    \centering
    \includegraphics[width=\linewidth]{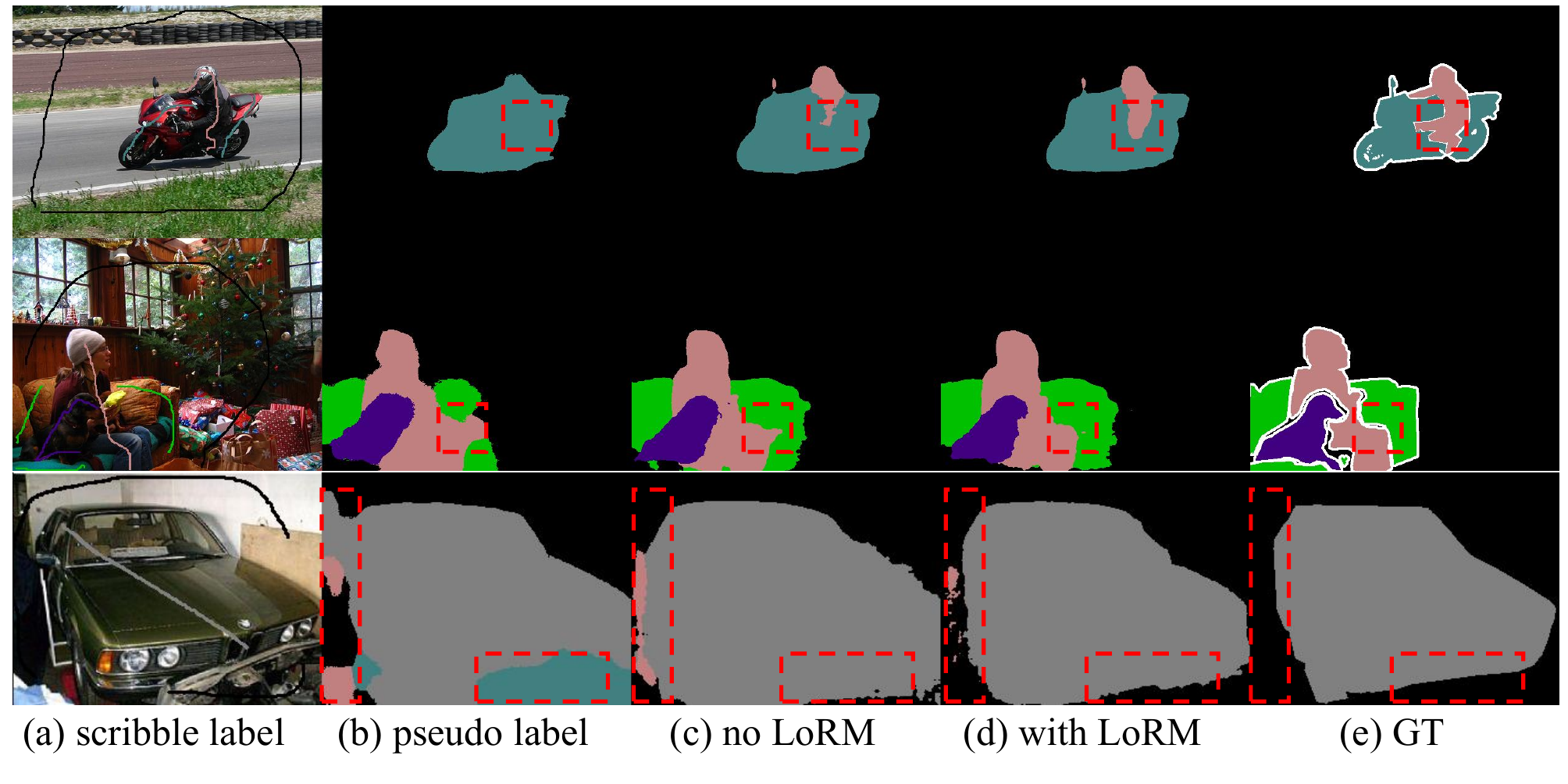}
    \caption{Visualization results employing resnet50 backbone and deeplabV2 segmentor. (a) is the original image with scribble label, (b) is the pseudo-label for training, (c) is the prediction trained with $\mathcal{L}_{basic}$, (d) is the prediction trained with $\mathcal{L}_{basic}+\mathcal{L}_{lorm}$. (e) is the ground truth label.}
    \label{fig:LoRM_vis}
\end{figure}
\begin{figure}
    \centering
    \includegraphics[width=\linewidth]{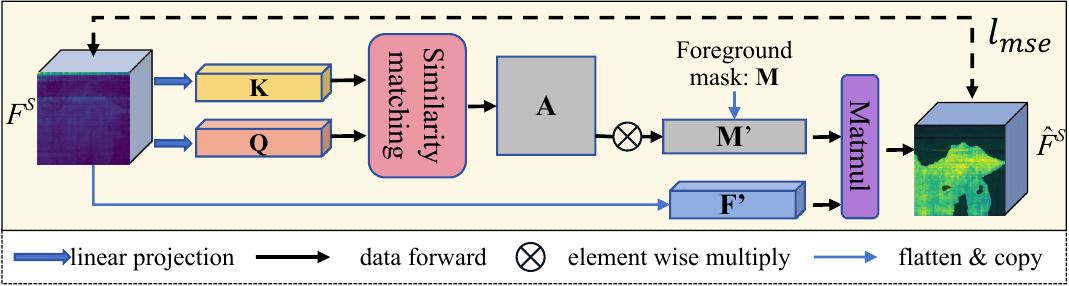}
    \caption{The workflow of the localization rectification module. ``Matmul" is short for the matrix multiplication. It takes the feature map $\mathbf{F}$ and the foreground pseudo mask $\mathbf{M}$ as inputs, and outputs the loss between the rectified feature map and the original feature map.}
    \label{fig:lorm}
\end{figure}
\begin{figure}
    \centering
    \includegraphics[width=\linewidth]{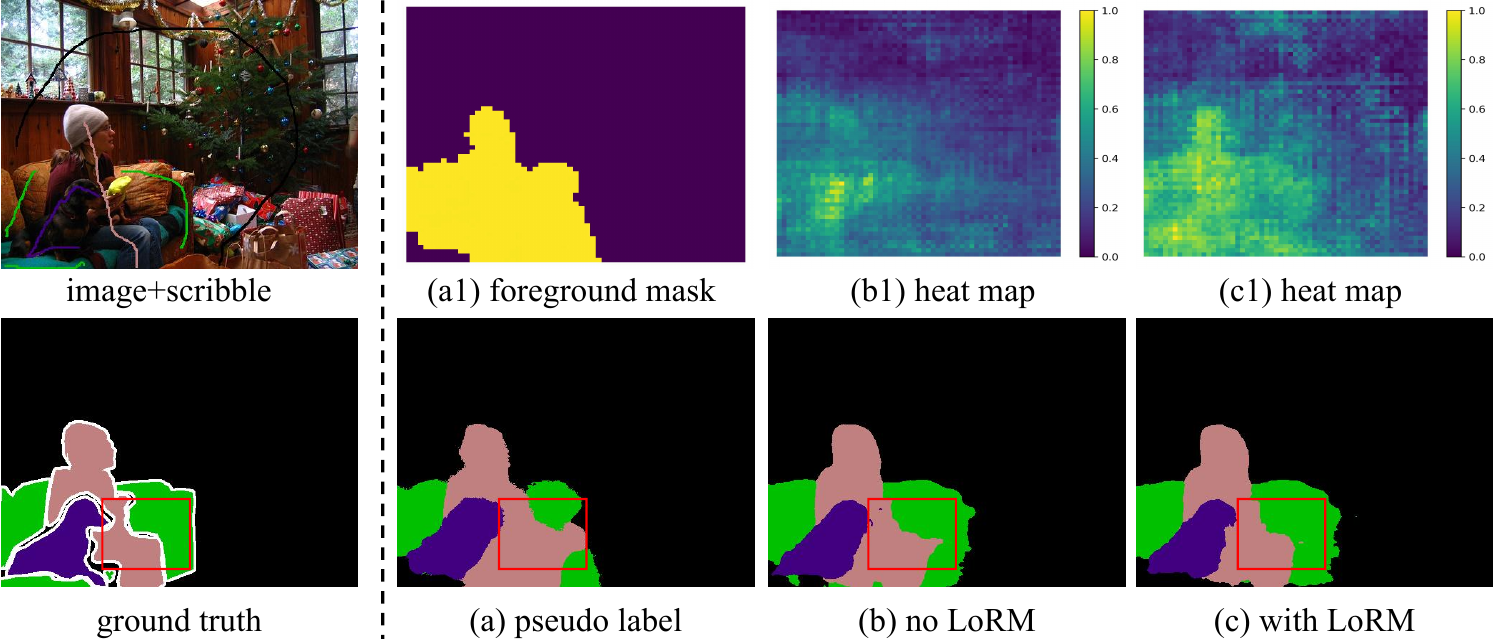}
    \caption{Feature map visualization of the last layer of resnet50 bachbone with deeplabV2 segmentator. (a) is the pseudo label and its corresponding foreground mask. (b) is the prediction without LoRM and its corresponding heat map. (c) is the prediction with LoRM and its corresponding heat map.}
    \label{fig:lormvisfeatures}
\end{figure}
\subsection{Localization rectification module}
The pseudo-labels generated in the first stage tend to confuse foreground objects when they are in close proximity, as shown in Fig. 3(b), which misleads the model into learning incorrect foreground objects. Therefore, it is crucial to prevent the model from overfitting to these noisy labels. To address this issue, we introduce the localization rectification module (LoRM)
Therefore, it is necessary to prevent the model from overfitting these noisy labels. To achieve this goal, we propose the localization rectification module (LoRM). The primary concept behind the LoRM is to leverage the inherent similarity of representations among foreground pixels belonging to the same semantic class. By doing so, mispredicted pixels can be refined through a weighted combination of representations from other pixels.  Let $\mathbf{F}^S \in \mathbb{R}^{C\times H \times W}$ denote the feature embeddings generated by the last layer of the segmentation backbone, and $\mathbf{M} \in \mathbb{R}^{H\times W}$ denotes the foreground binary mask, as depicted in Fig.~\ref{fig:cdsp}. The workflow of the LoRM is presented in Fig.~\ref{fig:lorm}

The feature embeddings $\mathbf{F}^S$ are firstly liner projected with a single convolution layer, then flattened along the row axis into $\mathbf{Q}\in \mathbb{R}^{C\times HW}$ and $ \mathbf{K} \in \mathbb{R}^{C\times HW}$. Taking $\mathbf{K}$ as the key set to be refined, and $\mathbf{Q}$ as the query set for similarity matching, we calculate the similarity matrix $\mathbf{A}$ by:
\begin{equation}
  \mathbf{A} = softmax(\frac{\mathbf{Q}^T \mathbf{K}}{\| \mathbf{Q}^T\Vert^{C}_2 \| \mathbf{K} \Vert^{C}_2}),
  \label{eq:similarity_A}
\end{equation}
where $\mathbf{A}\in \mathbb{R}^{HW\times HW}$, $softmax$ is implemented along the row axis, the L2-norm operation $\| \cdot \Vert_2^{C}$ of $\mathbf{Q}^T$ and $\mathbf{K}$ is implemented along the channel dimension. Each row vector $\mathbf{A}_i$ in the matrix $\mathbf{A}$ describes the similarity between the \emph{i-th} feature vector in $\mathbf{K}$ and all the $HW$ feature vectors in $\mathbf{Q}$. With the help of Eq.~\ref{eq:similarity_A}, the \emph{i-th} feature vector can be refined by referencing the feature vectors in other locations. It is worth noting that, the background vectors vary largely, and contribute little to the foreground rectification. Therefore, we extract the foreground mask $\mathbf{M}\in \mathbb{R}^{ H\times W}$ from the pseudo-label and flatten it along the row axis into $\mathbf{M'}$, then we element-wisely multiply it with $\mathbf{A}$ leveraging the broadcast technique:
\begin{equation}
  \mathbf{A'} = \mathbf{M'} \odot \mathbf{A},
\end{equation}
so that the background features in each row $\mathbf{A}_i$ are largely suppressed in its masked one $\mathbf{A'}_i$. Then the original feature map $\mathbf{F}^S$ is flattened along the row axis into $\mathbf{F'}^S$, and it is matrix-multiplied with the masked similarity matrix $\mathbf{A'}$:
\begin{equation}
  \mathbf{\hat{F}}^S = \delta * \mathbf{F'}^S \mathbf{A'} ,
\end{equation}
where $\delta$ is a learnable parameter initialized with 1 to control the rectification degree, $\mathbf{\hat{F}}^S \in \mathbb{R}^{C\times H W}$ is the refined feature embeddings, which is finally reshaped back to $\mathbb{R}^{C\times H\times W}$. The Mean Square Error loss (MSE) is implemented on the original feature $\mathbf{F}^S$ and the refined feature $\mathbf{\hat{F}}^S$:
\begin{equation}
  \mathcal{L}_{lorm} = \lambda_{lorm} * MSE(\mathbf{F}^S,\mathbf{\hat{F}}^S),
  \label{eq:l_LoRM}
\end{equation}
where $\lambda_{lorm}$ is a weight coefficient. The whole process is realized by efficient matrix operations. With the supervision of Eq.~\ref{eq:l_LoRM}, the LoRM achieves the goal of rectifying the misled foreground representations by referencing the representations in other foreground locations. Specifically, we present the heat map in Fig.~\ref{fig:lormvisfeatures} for a more intuitive illustration, where the feature map is a sum along the channel axis. From Fig.~\ref{fig:lormvisfeatures}(b) and (c), it can be seen that directly adopting the pseudo label will lead to the over-smooth representations, making the model hard to distinguish the foreground objects. In our experiments, we also found this module can significantly improve the model's stability, generating robust segmentation results with different scribble styles, which we will discuss in Section~\ref{sec:ablation}.

\subsection{Distance perception module}
The LoRM effectively addresses the misalignment in the feature space in the foreground area, but the model remains susceptible to be misled by noisy labels near the object boundary during later training steps. This could undermine the efforts of LoRM and reduce the model's robustness. 

To overcome this challenge, it becomes crucial to identify reliable predictions. We propose that discriminative areas, such as the surroundings of the scribble, are more reliable and should be assigned higher confidence. Conversely, indiscriminative areas like the boundary of the pseudo-label, generated by global class supervision, are less reliable and should be assigned lower confidence. Based on this concept, we introduce a distance perception strategy, to assign predictions with different confidence levels according to their distance from the scribble and the pseudo-label boundary respectively, named as the distance perception module. By doing so, we can better leverage the advantages of both supervisions during model training.

\begin{figure}[tbp]
  \centering
  \subfloat[Pseudo]{
      \includegraphics[width=0.22\linewidth]{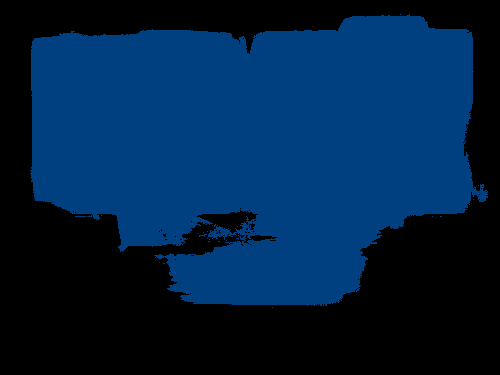}
      }
  \subfloat[$\lambda_c=1$]{
      \includegraphics[width=0.22\linewidth]{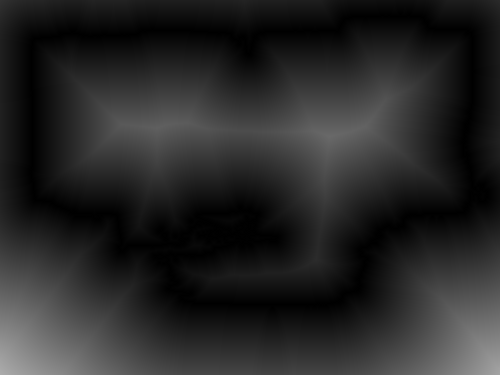}
      }
  \subfloat[$\lambda_c=3$]{
    \includegraphics[width=0.22\linewidth]{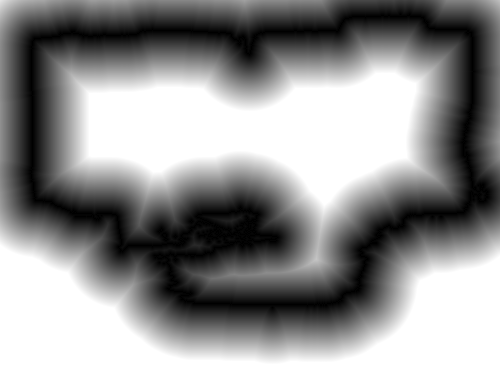}
      }
  \subfloat[$\lambda_c=7$]{
    \includegraphics[width=0.22\linewidth]{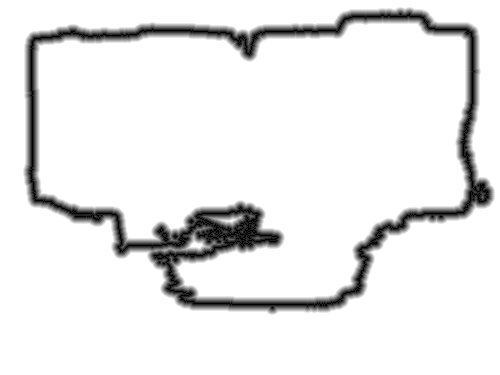}
      }
      
  \subfloat[Image]{
      \includegraphics[width=0.22\linewidth]{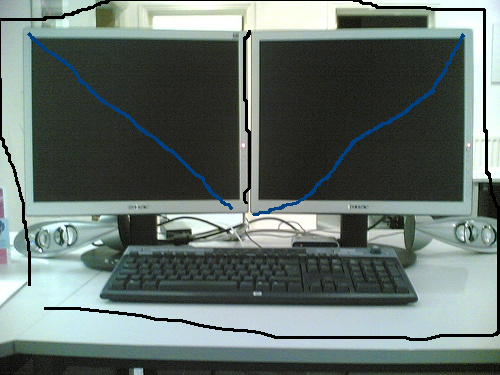}
      }
  \subfloat[$\lambda_s=1$]{
      \includegraphics[width=0.22\linewidth]{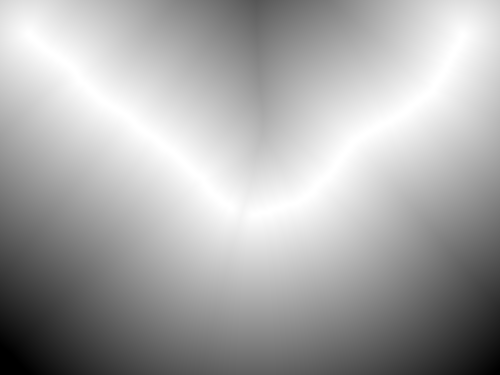}
      }
  \subfloat[$\lambda_s=3$]{
    \includegraphics[width=0.22\linewidth]{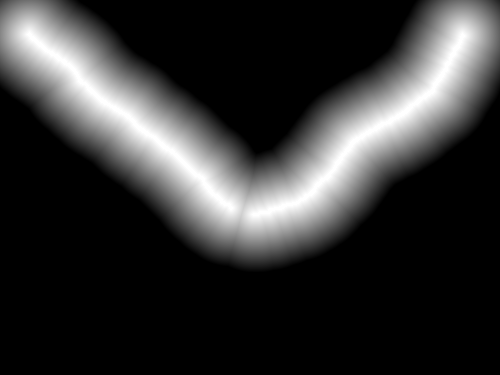}
      }
  \subfloat[$\lambda_s=7$]{
    \includegraphics[width=0.22\linewidth]{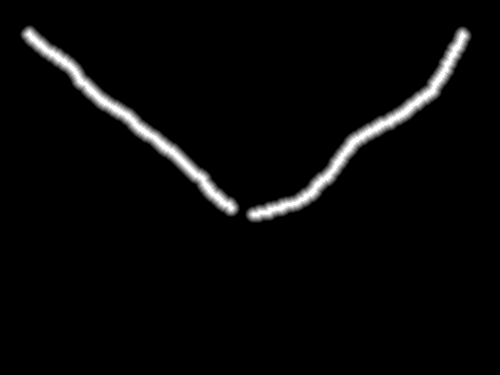}
      }

  \caption{ Visualization of distance maps with different coefficients for pseudo label boundary (b-d) and scribble (f-h)}
  \label{fig:DEM_vis}
\end{figure}

For the pseudo-label, the pixels around its boundary are indiscriminative, and such an area is probable to provide uncertain supervision. Denoting the coordinates of the $i^{th}$ point in the image as $(m,n)$, and the coordinates of the $j^{th}$ point on the foreground pseudo-label boundary as $(m',n')$, the distance maps of the pseudo-label is designed as:
\begin{equation}
  d_c(i)=\min\limits_{\forall j}(\frac{\lfloor \sqrt{e^{\lambda_c} [(m-m')^2+(n-n')^2]}\rfloor_{255}}{255}),
\end{equation}
where $d_c$ is a probability range in $[0,1]$ that describes the minimum Euclidean distance between a point and the set of pseudo-label boundary points with the distance value truncated to 255 for normalization and the efficiency of data storage. $\lambda_c$ is a coefficient to control the scope of the pseudo-label distance map as shown in Fig.~\ref{fig:DEM_vis} (f-h). Denoting $N_c$ as the number of non-zero elements in $d_c$, the distance perception of the pseudo-label is formulated as:
\begin{equation}
  \mathcal{L}_{dc} = \frac{1}{N_c} \sum_{i=1}^{N_c} d_c(i)\boldsymbol{p}_i log(\boldsymbol{p}_i).
  \label{eq:l_dc}
\end{equation}

Compared with the pseudo-label, the scribble is certain and correct, the pixels lying around the scribble may largely belong to the same semantic class as the scribble. Moreover, the scribble lying in the foreground's inner area provides correct supervision, which could suppress the noisy supervision in pseudo-label. But this confidence should decrease with the increment of the distance. Therefore, denoting the coordinates of the $i^{th}$ point in the image as $(m,n)$, and the $j^{th}$ foreground scribble point coordinates as $(m',n')$, the distance map of the scribble is designed as:
\begin{equation}
  d_s(i)=1-\min\limits_{\forall j}(\frac{\lfloor\sqrt{e^{\lambda_s} [(m-m')^2+(n-n')^2]}\rfloor_{255}}{255}),
\end{equation}
where $d_s$ is a probability range in $[0,1]$ that describes the minimum Euclidean distance between a point and the set of scribble points. $\lambda_s$ is a coefficient to control the scope of the scribble distance map as shown in Fig.~\ref{fig:DEM_vis}(b-d). Denoting $N_s$ as the number of nonzero elements in $d_s$, the distance perception of the scribble is formulated as:
\begin{equation}
  \mathcal{L}_{ds} = \frac{1}{N_s} \sum_{i=1}^{N_s} d_s(i) \boldsymbol{p}_i log(\boldsymbol{p}_i),
  \label{eq:l_ds}
\end{equation}
Finally, the overall distance perception can be formulated as:
\begin{equation}
  \mathcal{L}_{dp} = \mathcal{L}_{ds}+\mathcal{L}_{dc}.
  \label{eq:l_dem}
\end{equation}
Fig.~\ref{fig:DEM_vis} presents visualizations of the distance maps for the scribble and pseudo-label boundaries at different coefficients of $\lambda_s$ and $\lambda_c$. As $\lambda_s$ increases, the reliable area determined by the scribble becomes more prominent. Conversely, a higher $\lambda_c$ endows more weights to the pseudo-label in determining the reliable area. Through the distance perception module, we effectively excavate the reliable areas and reinforce the prediction certainty of the model by leveraging information from both the scribble and the pseudo-label boundaries.

\section{Scribble simulation algorithm}
\label{sec:scribble_simulation}
The current benchmark, ScribbleSup~\cite{lin2016scribblesup}, is built upon the Pascal VOC2012 dataset~\cite{pascal_voc}, consisting of 12,031 labeled images with only 21 classes. Given the advancements in segmentation methods within the weakly supervised semantic segmentation field, this dataset is insufficient for comprehensive evaluation and comparison. On the one hand, the dataset's relatively small size makes it prone to overfitting with modern models. On the other hand, the fixed scribble annotations in this dataset do not accurately reflect the variability in annotations that can occur with different human annotators, who may randomly label scribbles. As a result, the evaluation results on this dataset may not fully capture the real-world performance of methods, limiting the broader applicability and potential of scribble-supervised semantic segmentation.

\begin{figure}
    \centering
    \includegraphics[width=\linewidth]{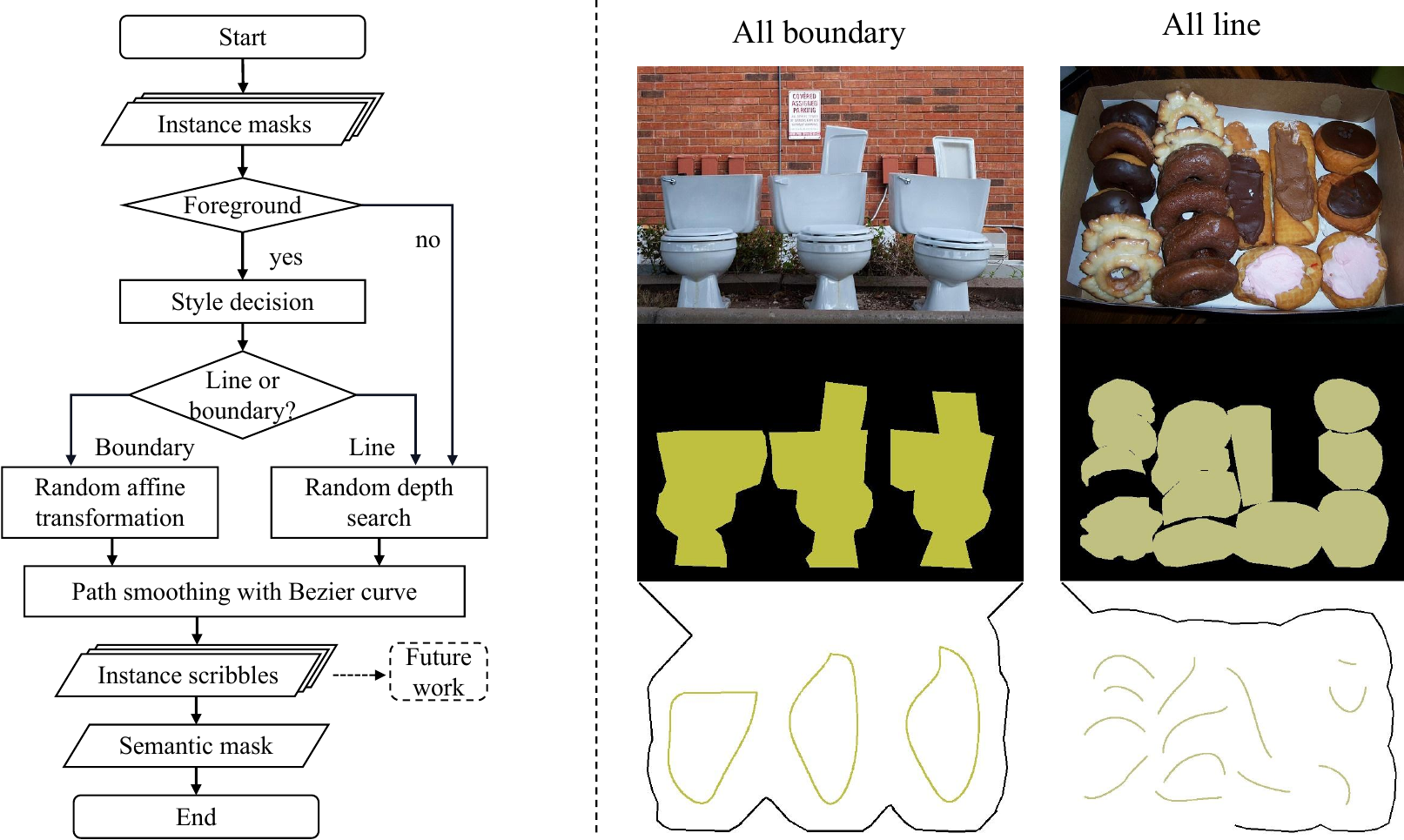}
    \caption{The workflow of our scribble simulation algorithm (left) and the demo of different scribble styles (right). The colorful scribble denotes foreground objects, the black scribble denotes the background.}
    \label{fig:scribble_simulation}
\end{figure}

Therefore, there is a need to develop a large-scale, scribble-annotated dataset for the scribble-supervised semantic segmentation field. To this end, we propose a scribble simulation algorithm that can be applied to existing instance-level annotated datasets, facilitating the creation of more diverse and realistic scribble annotations. The workflow is presented in Fig.~\ref{fig:scribble_simulation}. The codes will be made publicly available\footnote{https://github.com/Zxl19990529/ScribbleSimulation, code:``ayri"}.

\subsection{Style decision}

Given the instance annotation of an image, we first extract the background mask and the foreground object masks by instance. According to our observation in the ScribbleSup dataset, human annotators pretend to draw along the boundary for large objects and draw the curve in the object's center area, to achieve a balance in annotation accuracy and efficiency. Therefore, for each mask of the object, we decide the scribble style by calculating the proportion of the target object in the entire image:
\begin{equation}
    0< P = \frac{\text{The object pixel area}}{\text{The entire image pixel area}}<1,
\end{equation}
where the larger $P$ is, the more likely the boundary-style scribble will be annotated. Each scribble is generated in the format of $(x,y)$ point sequence, which is further connected and smoothed with the Bezier curve~\cite{beziercurve}. After that, all the scribbles are integrated into one semantic mask, which will be finally used for training the segmentation model.

\subsection{Random affine transformation}

\begin{figure}
    \centering
    \includegraphics[width=0.9\linewidth]{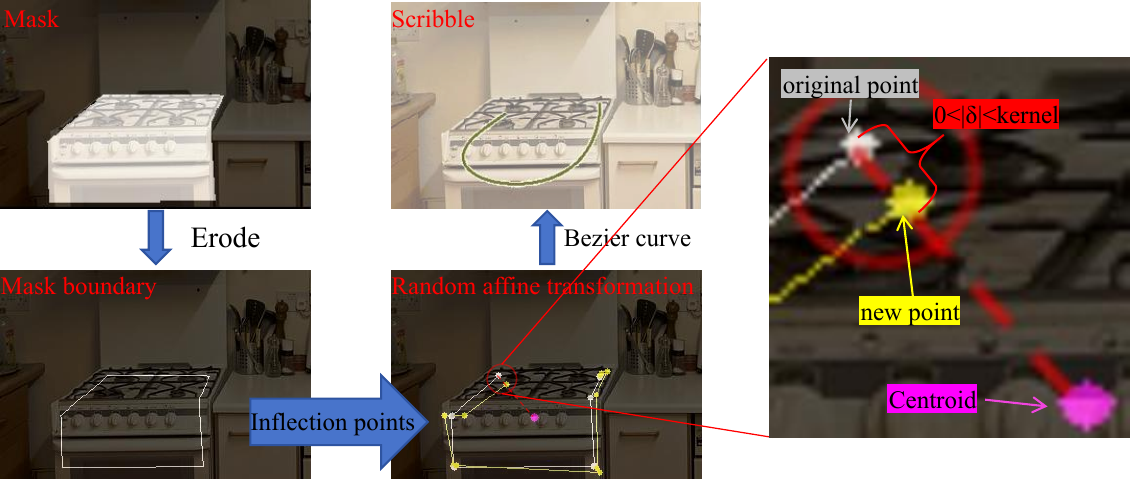}
    \caption{The illustration of the random affine transformation when generating boundary-style scribble. The white points are the boundary points extracted from the eroded mask, yellow points are the new points after the random affine transformation. The Bezier curve is implemented on the new points for connection and smoothing.}
    \label{fig:RandomAffine}
\end{figure}

This function aims to simulate the preference of drawing along the boundary of the object with the hand-shaking factor taken into consideration. Fig.~\ref{fig:RandomAffine} is presented to illustrate its workflow. For each instance, the input of this function includes three variables: the binary mask of the current object $M$, the kernel size for mask erosion $k$, and the random coefficient $\delta$. We first erode the $M$ with $k$ to make sure the scribble will appear inside the object, acquiring the eroded binary $M'$. Then we extract the boundary into an inflection point sequence $S=\{S_i=(x_i,y_i)| 0<i<n\}$. Denoting the centroid of $M'$ as $(c_x,c_y)$, each point of $S$ will be jittered using $\delta$:
\begin{equation}
\begin{aligned}
    &{x'}_i = x_i + \delta * cosine(arctan(\frac{y_i-c_y}{x_i-c_x}))\\
    &{y'}_i = y_i + \delta * sine(arctan(\frac{y_i-c_y}{x_i-c_x})),\\
\label{eq:random_affine}
\end{aligned}
\end{equation}
where $\delta$ is initialized before each calculation, ranging in $[-k,k]$. After the transformation, these points are organized into a new sequence $S'=\{{S'}_i=({x'}_i,{y'}_i)|0<i<n\}$, and will be sent to the Bezier curve for connection and smoothing.

\subsection{Random depth search}

This function aims to simulate the preference of drawing the curve in the object's center area. 
The pseudo-code in Python style is presented in Algorithm~\ref{alg:random_depth}. Given the binary mask of the current object $M$, it returns a random path $P$ chosen from the skeleton. \textbf{Step 1-3:} We first implement the medial axis transform on $M$ to acquire the skeleton $M'$, and build the skeleton into graph $G$ using MSP-Tracer~\cite{graphextract}. \textbf{Step 4-5:} However, the $G$ may contain multiple circle paths which may result in an endless loop when adopting the deep first search (DFS) algorithm~\cite{even2011graph}. Therefore, we further convert $G$ to the minimum spanning tree via the Kruskal algorithm~\cite{9357774} to remove the circle path. \textbf{Step 6-8:} The DFS will return a list of the paths, where each path is assigned with a probability according to the proportion of its length to the total length of all paths. The length is calculated using the Euclidean distance. \textbf{Step 9:} Finally, only one path $S$ will be selected according to the probability, which is a point sequence. After that, the path will be sent to the Bezier curve for connection and smoothing. Specifically, if the mask is background, it will not be smoothed to avoid covering the foreground.

\begin{algorithm}
\caption{Random depth search}
\label{alg:random_depth}
\KwIn{The binary mask of the current object $M$}
\KwOut{The selected path $S$}
Initialize the path list $L$, probability list $P$;\\
Implement medial axis transform on $M$ to acquire the skeleton $M'$;\\
Build the skeleton $M'$ into graph $G$; \\
Convert $G$ to Kruskal tree $T$;\\
Find all the paths in $T$: $L$= DFS($T$);\\
Calculate the sum of all path lengths $l$;\\
    \For{$L[i]$ in $L$}
    {
    P[i] = $\frac{length(L[i])}{l}$
    }
$S$= L[numpy.random.choice(P)]
\end{algorithm}

\begin{table}[h]
\centering
\caption{The dataset list.}
\begin{tabular}{lcccc} 
\toprule
\textit{\textbf{Previous datasets}}            & Train/val        & Resolution$^{*}$       & \makecell[c]{Classes}  \\ 
\hline
ScribbleSup~\cite{lin2016scribblesup}        & 10582/1449       & (88,400)-(500,500)      & 21                              \\
ScribbleACDC~\cite{olivier2018acdc}       & 1711/191                  & (154,224)-(512,428)    & 4                           \\
\midrule
\multicolumn{4}{l}{\textit{\textbf{New datasets}}}\\
\hline
ScribbleCOCO       & 82081/40137      & (51,72)-(640,640)        & 81                                  \\
ScribbleCityscapes & 2965/472         & (1024,2048)            & 9                                   \\
\bottomrule
\multicolumn{4}{l}{$*$\footnotesize formatted in (height,width).}
\end{tabular}
\label{tab:datasets}
\end{table}

\subsection{Dataset}
In this paper, we proposed two large-scale datasets including ScribbleCOCO and ScribbleCityscapes, where the scribble annotations of these two datasets were generated with our scribble simulation algorithm three times, for evaluating the robustness and accuracy of different methods. We also reported the accuracy of our method on public human-annotated scribble datasets, including ScribbleSup and ScribbleACDC. The comparison between our newly proposed datasets and the previous datasets is presented in Table.~\ref{tab:datasets}. 

\textbf{ScribbleCOCO} is a new large-scale dataset which is made based on MicroSoft Common Objects in Context~\cite{lin2014microsoft} with our proposed scribble simulation algorithm in Section~\ref{sec:scribble_simulation}. The kernel size $k$ for erosion in Eq.~\ref{eq:random_affine} is set to 20. This dataset contains 82081 training images and 40137 validation images with 80 foreground object classes and 1 background class. Some examples of the ScribbleCOCO are presented in Fig.~\ref{fig:ScribbleCOCO_vis}.

\textbf{ScribbleCityscapes} is a new large-scale urban scene dataset which is made based on Cityscapes~\cite{cordts2016cityscapes} with our proposed scribble simulation algorithm in Section~\ref{sec:scribble_simulation}. The kernel size $k$ for erosion in Eq.~\ref{eq:random_affine} is set to 20. This dataset contains 2965 training images and 472 validation images with 8 foreground object classes and 1 background class. All the images have the same resolution of 2048 pixels width and 1024 pixels height. Specifically, as the images are captured inside the front windshield, we cut the image into two parts horizontally when generating the background scribble. Some examples are presented in Fig.~\ref{fig:ScribbleCityscapes_vis}.

\textbf{ScribbleSup} is a public dataset which combined PASCAL VOC2012 and SBD~\cite{hariharan2011semantic} datasets with one style of the scribble annotations~\cite{lin2016scribblesup}. The dataset includes 10,582 training images and 1,449 validation images with 20 foreground object classes and 1 background class. Additionally, we validated our method on \emph{scribble-shrink} and \emph{scribble-drop} introduced by URSS~\cite{pan2021scribble} to assess its robustness in diverse scenarios.

\textbf{ScribbleACDC} is a public 3D medical image dataset, which consists of 100-MRI scans with manual scribble annotations for RV, LV, and MYO provided by Valvano et al.~\cite{valvano2021learning}. In this paper, we cut each 3D MRI volume into slices, and randomly split these slices with a train-val ratio of $9:1$, acquiring 1711 slices for training and 191 slices for validation.

\begin{figure}
    \centering
    \includegraphics[width=\linewidth]{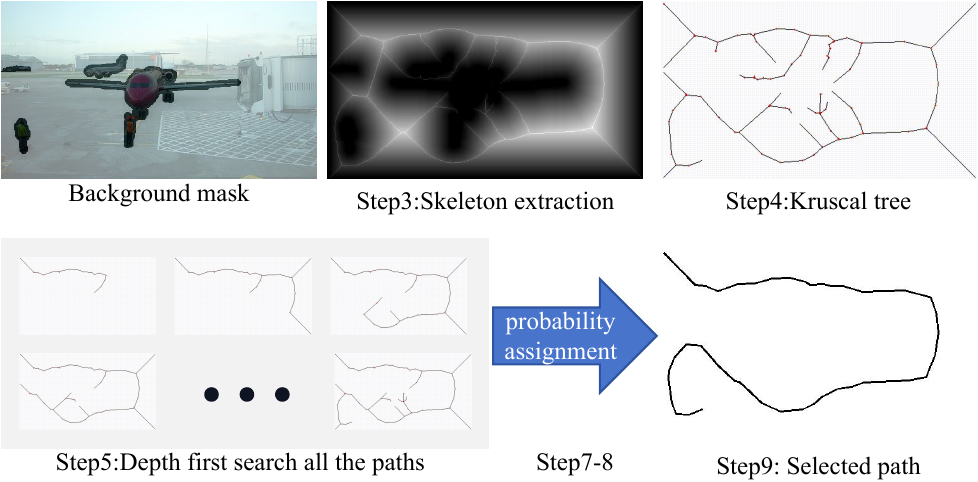}
    \caption{The illustration of the random depth search together with Algorithm~\ref{alg:random_depth}. The background mask is taken as an example here. The example image is taken from VOC2012.}
    \label{fig:RandomDepthSearch}
\end{figure}

\begin{figure*}
    \centering
    \includegraphics[width=\linewidth]{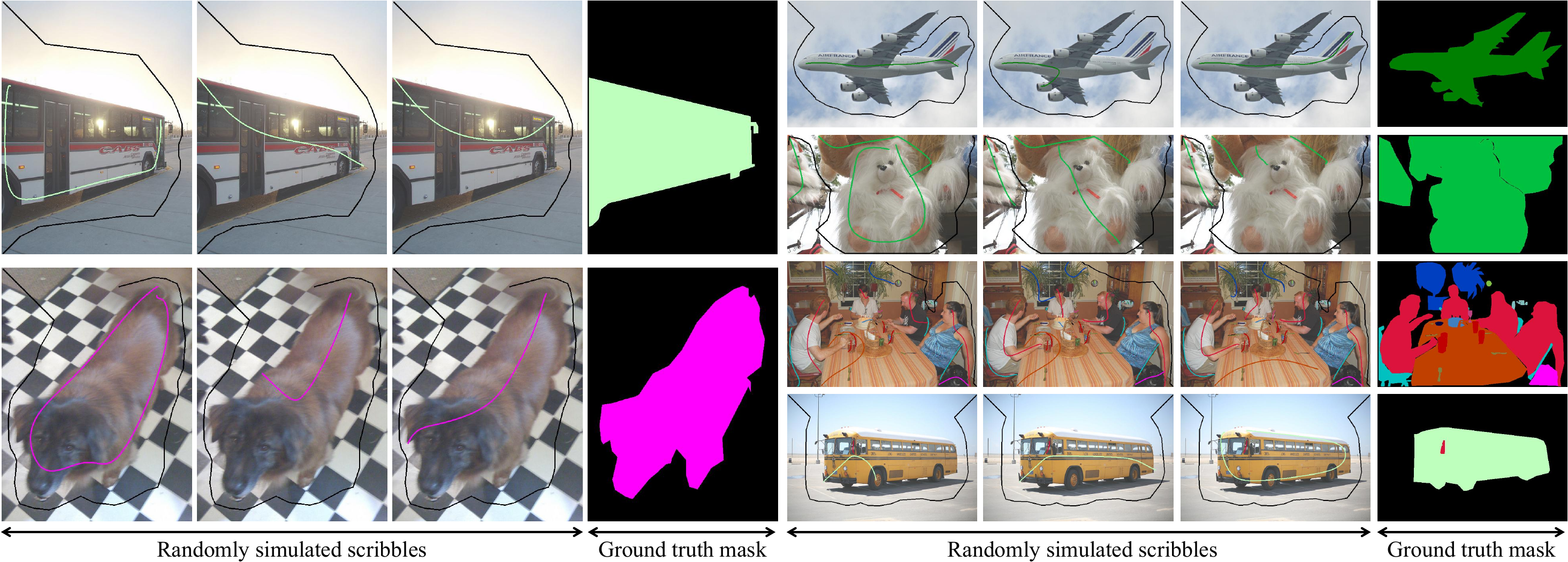}
    \caption{Some examples of the proposed ScribbleCOCO. The scribbles are generated with our simulation algorithm, where the human annotators' preferences and randomness are taken into consideration. We generated three different scribbles in this dataset and conducted the experiments three times to evaluate the model's performance. The black scribble represents the background, other colors represent the foreground objects.}
    \label{fig:ScribbleCOCO_vis}
\end{figure*}

\begin{figure*}
    \centering
    \includegraphics[width=\linewidth]{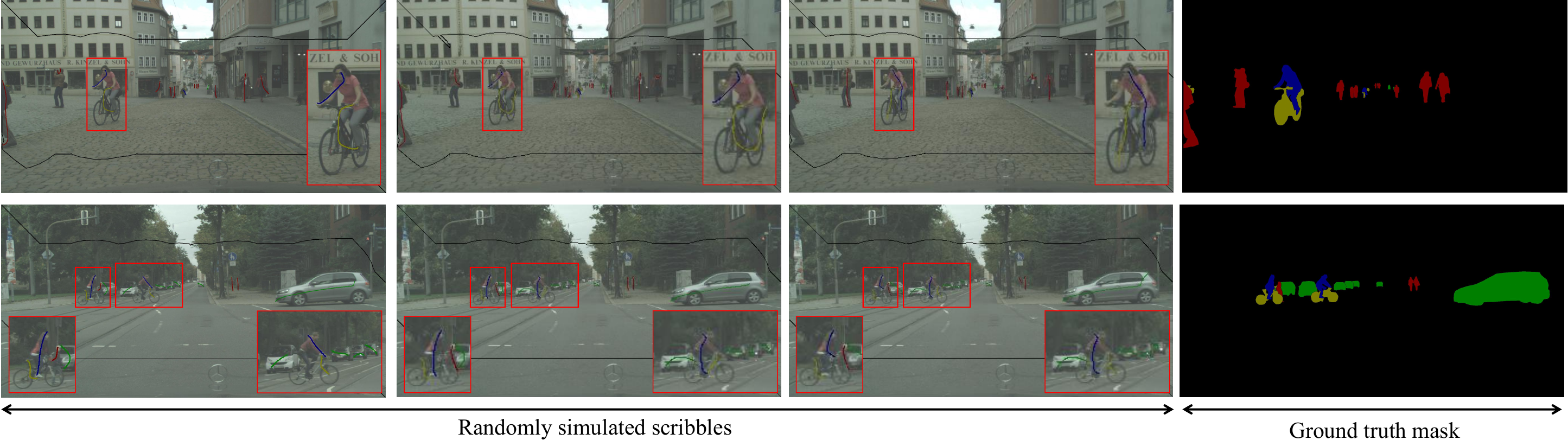}
    \caption{Some examples of the proposed ScribbleCityscapes. The scribbles are generated with our simulation algorithm, where the human annotators' preferences and randomness are taken into consideration. As the foreground objects are small, we present specific regions with a zoomed-in view. The black scribble represents the background, other colors represent the foreground objects}
    \label{fig:ScribbleCityscapes_vis}
\end{figure*}
\section{Experiments}

\subsection{Implementation details}

In general, our method was implemented in a pure ``Python + PyTorch 2.0" environment. All experiments were conducted on a platform equipped with NVIDIA Tesla L40S GPUs. For the reproduction of other methods, we used the official codebases provided by the respective repositories.

\textbf{Stage 1}: In this stage, we utilized the Distributed Data Parallel (DDP) function in PyTorch to train the scribble-promoted ToCo with Eq.~\ref{eq:l_toco} loss function. For different datasets, the training hyper-parameters are listed in Table.~\ref{tab:toco_parameter} in the appendix. The vision transformer (ViT)~\cite{dosovitskiy2020image} was adopted as the backbone. Other settings were kept unchanged with the original ToCo. After training, we infer the pseudo-labels on the train sets with the last checkpoint.

\textbf{Stage 2}: In this stage, we employed the DeeplabV3$+$~\cite{chen2018encoder} as the segmentor with resnet101~\cite{he2016deep} as the backbone. For all the datasets, we conducted a total of 50 epochs with the Stochastic Gradient Descent (SGD) optimizer, where the momentum was set to 0.9. To ensure stable training, we implemented the linear warmup strategy to the learning rate for the first 10 epochs, and cosine annealing the learning rate to zero for the next 40 epochs. The batch size was set to 16 for all the datasets. All the experiments in this stage were conducted on one GPU, and the last checkpoint was used for inference. Other training hyper-parameters for each dataset were listed in Table.~\ref{tab:cdsp_params} in the appendix.

In all the experiments, the ImageNet-pretrained~\cite{jia2009imgnet} weights were utilized to initialize the model in terms of the backbones~\cite{he2016deep,dosovitskiy2020image}.

\section{Results and analysis}
\subsection{Quantitative results}

\textbf{Results on ScribbleCOCO and ScribbleCityscapes}: On these datasets, we generated the scribble annotations 3 times with our scribble simulation algorithm to acquire random scribble styles.  The experiments were conducted 3 times with the same hyper-parameters, which are denoted as ``s1" (short for scribble 1), ``s2", and ``s3".  We reimplemented the recent milestones including URSS~\cite{xu2021scribble}, TEL~\cite{liang2022tree}, and AGMM~\cite{wu2023sparsely} on these datasets. The baseline method was a DeeplabV3$+$ model with resnet101 backbone trained on scribble exclusively with partial cross-entropy loss. The comparison results are recorded in Table.~\ref{tab:comparison_coco2014} and Table.~\ref{tab:comparison_cityscapes} respectively. From the table, our CDSP achieves the state-of-the-art performance on both accuracy and robustness.

\begin{table*}
\centering
\caption{Comparison on ScribbleCOCO2014 (mIoU).$\uparrow$ means higher is better, $\downarrow$ means lower is better }
\begin{tabular}{lp{1.3cm}<{\centering}p{1cm}<{\centering}p{1.5cm}<{\centering}p{1.1cm}<{\centering}p{1.1cm}<{\centering}p{1.1cm}<{\centering}p{1.1cm}<{\centering}p{1.1cm}<{\centering}} 
\toprule
Method      & Pub                             & \textit{Sup} & Segmentor & s1      & s2      & s3      & mean$\uparrow$    & std$\downarrow$     \\ 
\hline
\multicolumn{2}{l}{Fully supervised}         &$\mathcal{F}$   & v3p+r101  & \multicolumn{3}{c}{-}       & 55.73\% & -       \\
\multicolumn{2}{l}{Fully supervised}       &$\mathcal{F}$   & v3p+r50   & \multicolumn{3}{c}{-}       & 53.74\% & -       \\
\hline
HAS~\cite{kumar2017hide}        & 17'ICCV    & $\mathcal{I}$   & v3p+r101  & \multicolumn{3}{c}{-} & 38.51\% & -    \\
Cutmix~\cite{yun2019cutmix}     & 19'ICCV    & $\mathcal{I}$   & v3p+r101  & \multicolumn{3}{c}{-} & 38.43\% & -    \\
ReCAM~\cite{chen2022class}      & 22'CVPR    &$\mathcal{I}$   & v3p+r101  & \multicolumn{3}{c}{-}       & 39.54\% & -       \\
ToCo~\cite{ru2023token}         & 23'CVPR    &$\mathcal{I}$   & ToCo      & \multicolumn{3}{c}{-}       & 42.30\% & -       \\ 
ToCo*(ours)                     & -          &$\mathcal{I+S}$   & ToCo    & 45.22\% &  45.18\% &  45.29\%&  45.23\% & 0.05\% \\ 
\hline
\multicolumn{2}{l}{baseline(scribble only)}   &$\mathcal{S}$   & v3p+r101  & 52.37\% & 52.32\% & 52.18\% & 52.29\% & 0.10\%  \\ 
URSS~\cite{xu2021scribble}      & 21'ICCV    & $\mathcal{S}$   & v2+r101   & 51.13\% & 50.67\% & 51.20\% & 51.00\% & 0.29\%  \\
TEL~\cite{liang2022tree}        & 22'CVPR    &$\mathcal{S}$   & v3p+r101  & 52.84\% & 52.77\% & 52.74\% & 52.78\% & 0.05\%      \\
AGMM~\cite{wu2023sparsely}      & 23'CVPR    &$\mathcal{S}$   & v3p+r101  & 54.12\% & 52.23\% & 54.23\% & 53.52\% & 1.12\%  \\
\hline
Ours(HAS) & -                               &$\mathcal{S}$   & v3p+r101   & 54.09\% & 54.07\% & 54.01\% & 54.06\%  & \textbf{0.04}\%  \\
Ours(Cutmix) & -                            &$\mathcal{S}$   & v3p+r101  & 54.06\%  & 54.09\% & 53.97\% & 54.06\% & 0.07\%    \\
Ours(ReCAM) & -                             &$\mathcal{S}$   & v3p+r101  & 54.49\%  & 54.34\% & 54.30\% & 54.38\% & 0.10\%    \\
Ours(ToCo*)  & -                            &$\mathcal{S}$   & v3p+r50   & 53.05\% & 53.17\% & 52.93\% & 53.05\% & 0.12\%  \\
Ours(ToCo*) & -                             &$\mathcal{S}$   & v3p+r101  & \textbf{54.49}\% & \textbf{54.64}\% & \textbf{54.37}\% & \textbf{54.50}\% & 0.14\%  \\

\bottomrule
\multicolumn{9}{p{14.5cm}}{*ToCo is improved with the scribble label during the training phase in stage 1. $\mathcal{I}$: supervised by the image-level class label. $\mathcal{S}$: supervised by scribble label. $\mathcal{F}$: fully supervised. Results of s1, s2, and s3 are trained with different styles of scribble label.}
\end{tabular}
\label{tab:comparison_coco2014}
\end{table*}
\begin{table*}
\centering
\caption{Comparison on ScribbleCityscapes (mIoU).$\uparrow$ means higher is better, $\downarrow$ means lower is better }

\begin{tabular}{lp{1.3cm}<{\centering}p{1cm}<{\centering}p{1.5cm}<{\centering}p{1.1cm}<{\centering}p{1.1cm}<{\centering}p{1.1cm}<{\centering}p{1.1cm}<{\centering}p{1.1cm}<{\centering}} 
\toprule
Method      & Pub                             & \textit{Sup} & Segmentor & s1      & s2      & s3      & mean$\uparrow$    & std$\downarrow$     \\ 
\hline 
\multicolumn{2}{l}{Fully supervised}   &$\mathcal{F}$& v3p+r101  & \multicolumn{3}{c}{-}                                                                   & 67.26\%                  & \multicolumn{1}{c}{-}    \\
\multicolumn{2}{l}{Fully supervised}  &$\mathcal{F}$& v3p+r50   & \multicolumn{3}{c}{-}                                                                   & 67.64\%                  & \multicolumn{1}{c}{-}    \\
\hline
\multicolumn{2}{l}{baseline(scribble only)} &$\mathcal{S}$    & v3p+r101  & 52.73\%                     & \multicolumn{1}{c}{52.55\%} & \multicolumn{1}{c}{52.53\%} & 52.60\%                  & \textbf{0.11}\%                   \\

URSS~\cite{xu2021scribble}       & 21'ICCV   &$\mathcal{S}$    & v2+r101   & \multicolumn{1}{c}{44.88\%} & 44.55\%                     & 47.50\%                     & 45.65\%                  & 1.62\%                   \\
TEL~\cite{liang2022tree}        & 22'CVPR   &$\mathcal{S}$    & v3p+r101  & 53.55\%                     & 53.70\%                     & 53.39\%                     & 53.54\%                  & 0.15\%                   \\
AGMM~\cite{wu2023sparsely}       & 23'CVPR   &$\mathcal{S}$    & v3p+r101  & 46.71\%                     & 48.37\%                     & 50.14\%                     & 48.41\%                  & 1.72\%                   \\
Ours       &    -     &$\mathcal{S}$    & v3p+r50   & 57.33\%                     & \multicolumn{1}{c}{57.53\%} & \multicolumn{1}{c}{57.32\%} & 57.39\%                  & 0.12\%                   \\
Ours       &    -     &$\mathcal{S}$    & v3p+r101  & \multicolumn{1}{c}{\textbf{59.46}\%} & \textbf{58.24}\%                     & \textbf{59.03}\%                     & \textbf{58.91}\%                  & 0.62\%                   \\

\bottomrule
\end{tabular}
\label{tab:comparison_cityscapes}
\end{table*}
In Table.~\ref{tab:comparison_coco2014}, URSS is a pseudo-label free method, which aims to improve the model's certainty with the help of consistency loss and minimum entropy loss. However, as the ScribbleCOCO is much larger than ScribbleSup, obstinately encouraging the consistency and certainty of the model may make the model overconfident, which makes the model scribble-sensitive and hinders the model's robustness. TEL is a typical self-training based method, which utilizes the model's prediction as the pseudo label. Though TEL is stable in the three experiments, it achieves a very limited segmentation performance with an average mIoU of only 52.78\%, reflecting the typical Matthew Effect that the model overfits the preferred predictions generated by itself. AGMM utilized a Gaussian learner to adaptively refine the model's prediction as the pseudo label. However, the randomness of the scribble does not follow the Gaussian process, as the annotators have their inherent preference for labeling. Furthermore, the Gaussian prior is not always suitable for all the datasets. Therefore, AGMM reflects an unstable segmentation performance, where it generates competitive results on the ``s1'' and ``s3'' scribble with 54.12\%  and 54.23\% mIoU scores respectively, it performs unsatisfactory results on the ``s2'' scribble with only 52.23\%.

\begin{figure*}
    \includegraphics[width=\linewidth]{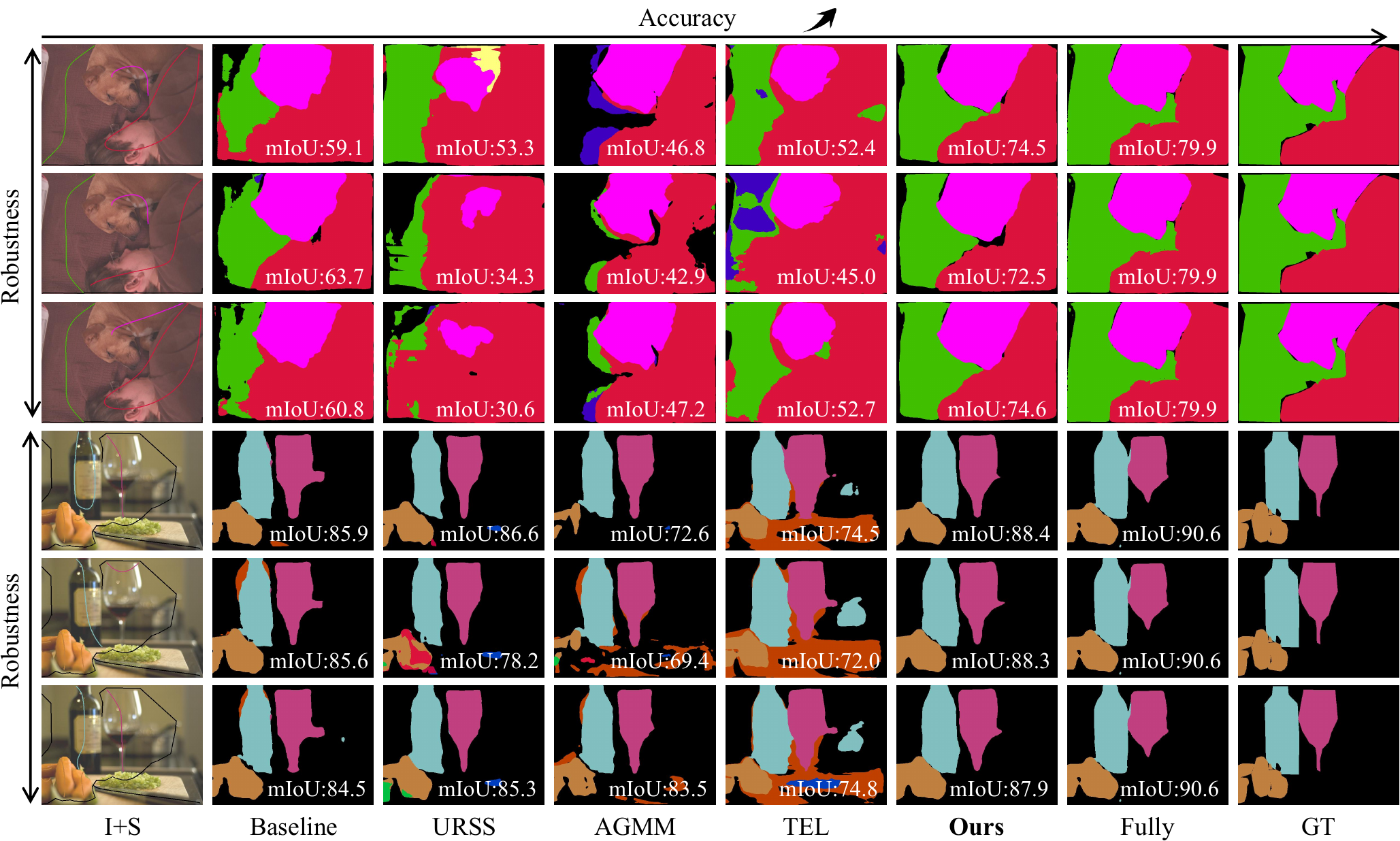}
    \caption{Visual comparison of different methods on the ScribbleCOCO train set. From the left to right reflects the model's accuracy, and from the top to bottom reflects the model's robustness and stability. The baseline is a deeplabv3p model with resnet-101 backbone trained with only the scribble label. URSS is scribble-sensitive due to the lack of the pseudo label. AGMM fails to make accurate predictions as Gaussian prior is unsuitable for all the scribbles. TEL easily falls into the Matthew Effect due to its self-training strategy.}
    \label{fig:comparecoco-train}
\end{figure*}

In Table.~\ref{tab:comparison_cityscapes}, our method achieved state-of-the-art performance on all three scribble styles, which significantly outperforms other methods with the average mIoU of 57.39\% and 0.12\% standard deviation with resnet-50 backbone. With the resnet-101 backbone, the performance is even better, with a higher mean mIoU of 58.91\%, exceeding AGMM by 10.5\%, TEL by 5.37\%, and URSS by 13.26\%. It has to be noted that, the ScribbleCityscapes is a high-resolution dataset with the image size of 2048 width and 1024 height. From the results, the lack of pseudo-label supervision becomes the weakness of URSS. TEL, though insensitive to the scribble shape, suffers from the Matthew Effect, which limits the model's potential. AGMM has the highest standard deviation of 1.72\%, where the Gaussian prior failed to adaptively refine the model's prediction on this urban scene dataset, showcasing its unrobustness to high-resolution images. 
\begin{figure*}
    \includegraphics[width=\linewidth]{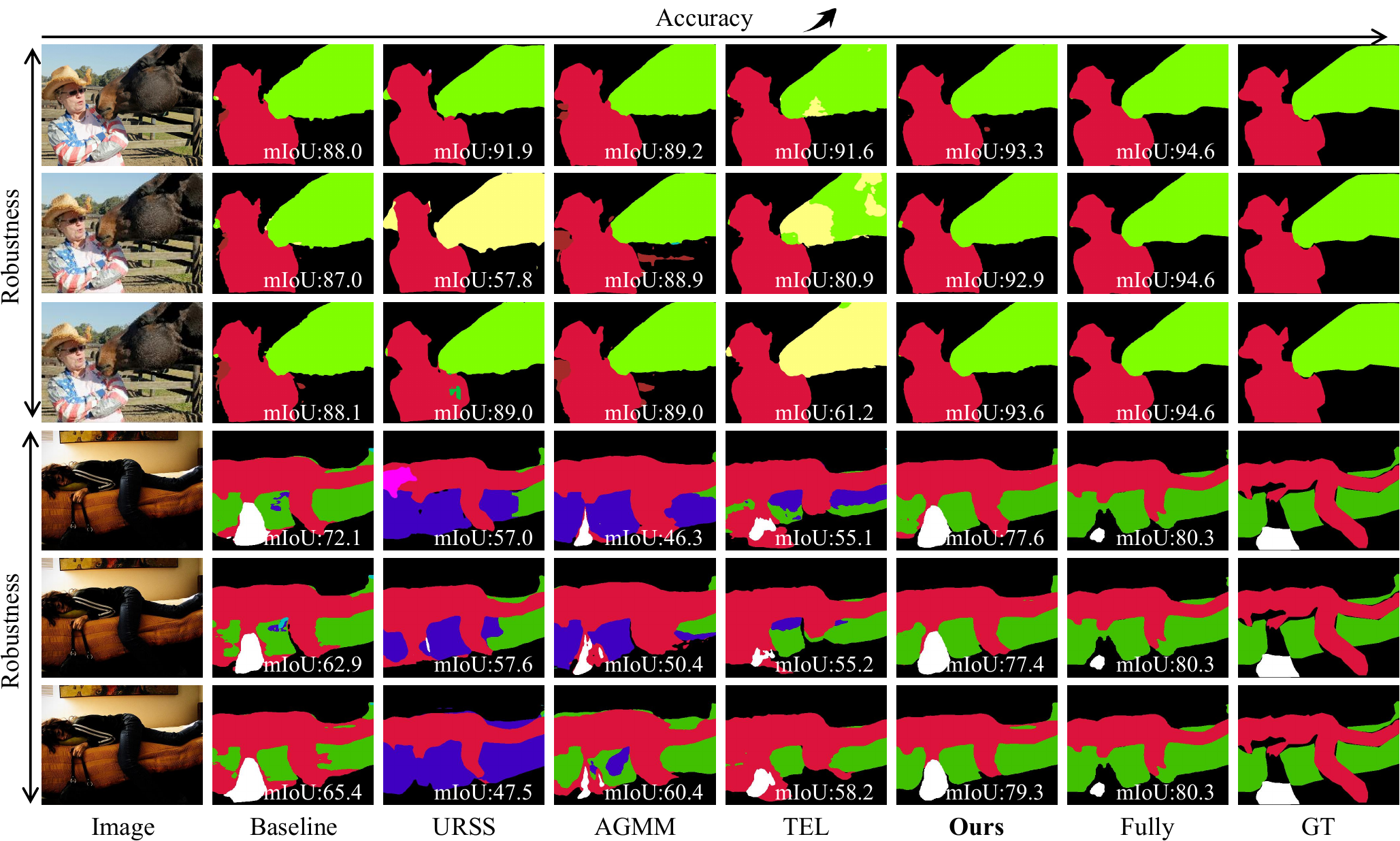}
    \caption{Visual comparison of different methods on the ScribbleCOCO val set. From the left to right reflects the model's accuracy, from the top to bottom reflects the model's robustness and stability. The baseline is a DeeplabV3$+$ model with resnet-101 backbone trained with only the scribble label. URSS is scribble sensitive due to the lack of the pseudo label. AGMM fails to make accurate predictions as gaussian prior is not suitable to all the scribbles. TEL is easy to fall into Matthew Effect due to its self-training strategy.}
    \label{fig:comparecoco-val}
\end{figure*}
\begin{figure*}
    \includegraphics[width=\linewidth]{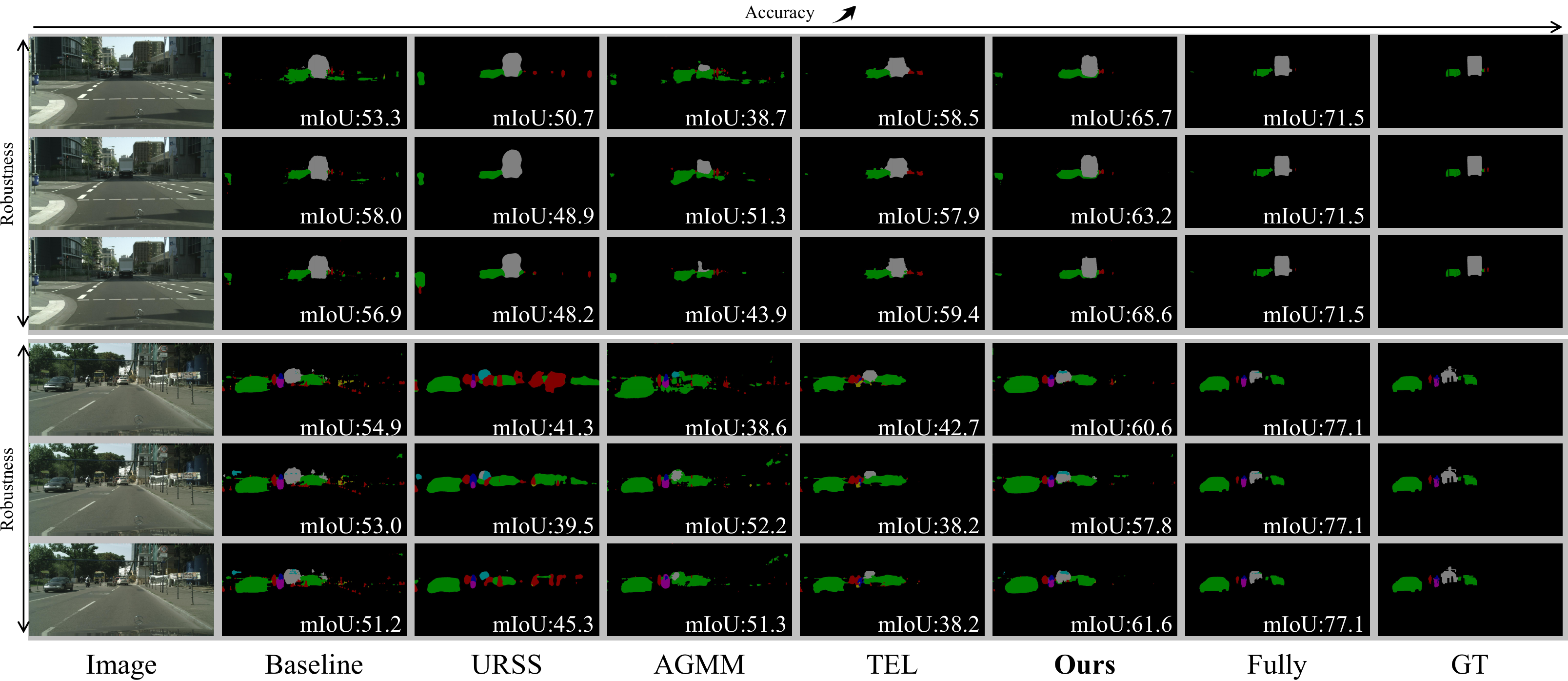}
    \caption{Prediction visualization of different methods on the ScribbleCityscapes val set. Our method outperforms other milestones regarding segmentation accuracy while maintaining robustness. Better view with zoomed-in.}
    \label{fig:comparecityscapes-val}
\end{figure*}

In comparison, our method is not only accurate and robust on the common resolution dataset like ScribbleCOCO, but also appealing on the high-resolution dataset like ScribbleCityscapes.
\begin{table}
\centering
\caption{Comparison on ScribbleSup (mIoU).$\uparrow$means higher is better}
\begin{tabularx}{\linewidth}{lccccc} 
\toprule
Method      & Pub        & \textit{Sup} & Segmentor & mIoU$\uparrow$ & CRF    \\ 
\hline
\multicolumn{2}{l}{Fully supervised}             & $\mathcal{F}$           & v3p+r101  & 78.14\% & - \\
\multicolumn{2}{l}{Fully supervised}            & $\mathcal{F}$           & v3p+r50   & 75.88\% & - \\
\hline
SEAM~\cite{wang2020self}        & 20'CVPR     & $\mathcal{I}$           & v1+r38    & 64.50\%  & -\\
AFA~\cite{zhang2021affinity}         & 22'CVPR     & $\mathcal{I}$          & SegFormer & 66.00\% & - \\
BMP~\cite{zhu2023branches}         & 24'TNNLS     & $\mathcal{I}$          & BMP       & 68.10\% & - \\
ToCo~\cite{ru2023token}        & 23'CVPR     & $\mathcal{I}$          & ToCo      & 71.10\% & - \\ 
\multicolumn{2}{l}{ToCo*(ours)}     & $\mathcal{I+S}$          & ToCo      & \textbf{72.72}\% & - \\ 
\hline
ScribbleSup~\cite{lin2016scribblesup} & 16'CVPR   & $\mathcal{S}$           & v1+vgg16  & 63.10\% & \checkmark \\
RAWKS~\cite{vernaza2017learning}       & 17'CVPR   & $\mathcal{S}$           & v1+r101   & 61.40\% & \checkmark \\
KCL~\cite{tang2018regularized}         & 18'ECCV   & $\mathcal{S}$           & v2+r101   & 72.90\% & \checkmark \\
NCL~\cite{tang2018normalized}         & 18'CVPR   & $\mathcal{S}$           & v1+r101   & 72.80\% & \checkmark \\
BPG ~\cite{wang2019boundary}        & 19'IJCAI    & $\mathcal{S}$           & v2+r101   & 73.20\% & - \\
URSS~\cite{pan2021scribble}        & 21'ICCV     & $\mathcal{S}$           & v2+r101   & 74.60\% & - \\
PSI~\cite{xu2021scribble}         & 21'ICCV     & $\mathcal{S}$           & v3p+r101  & 74.90\% & - \\
CCL~\cite{wang2022cycle}         & 22'HCMA & $\mathcal{S}$           & v2+r101   & 74.40\% & - \\
PCE~\cite{li2022weakly}         & 22'NPL      & $\mathcal{S}$           & v3p+r101  & 73.80\% & - \\
TEL~\cite{liang2022tree}         & 22'CVPR     & $\mathcal{S}$           & v3p+r101  & 75.23\% & - \\
AGMM~\cite{wu2023sparsely}        & 23'CVPR     & $\mathcal{S}$           & v3p+r101  & 74.24\% & - \\ 
\multicolumn{2}{l}{baseline(scribble only)}         & $\mathcal{S}$           & v2+r50  & 66.17\% & - \\ 
\hline
Ours(SEAM)  & -          & $\mathcal{S}$           & v3p+r101  & 71.77\% & - \\
Ours(AFA)   & -          & $\mathcal{S}$           & v3p+r101  & 73.31\% & - \\
Ours(BMP)   & -          & $\mathcal{S}$           & v2+r50  & 73.92\% & - \\
Ours(BMP)   & -          & $\mathcal{S}$           & v2+r101  & 75.25\% & - \\
Ours(BMP)   & -          & $\mathcal{S}$           & v3p+r101  & 75.85\% & - \\
Ours(ToCo*)  & -          & $\mathcal{S}$           & v3p+r101  & \textbf{75.86}\% & - \\ 
\bottomrule
\multicolumn{6}{X}{*ToCo is improved with the scribble label during the training phase in stage 1. $\mathcal{I}$: supervised by the image-level class label. $\mathcal{S}$: supervised by scribble label. $\mathcal{F}$: fully supervised.}
\end{tabularx}
\label{tab:comparison_scribblesup}
\end{table}

\begin{table}[htbp]
\centering
\caption{Comparison on ScribbleACDC (mIoU).$\uparrow$ means higher is better, $\downarrow$ means lower is better }
\begin{tabular}{lcccc} 
\toprule
Method      & Pub        & \textit{Sup} & Segmentor & mIoU$\uparrow$ \\ 
\hline
\multicolumn{2}{l}{Fully supervised}  &$\mathcal{F}$   & v3p+r50    & 89.70\%  \\
\multicolumn{2}{l}{Fully supervised}  &$\mathcal{F}$   & v3p+r101    & 89.36\%  \\
\hline
UNet$*$~\cite{ronneberger2015u}(baseline)        & 15'MICCAI &$\mathcal{S}$          & Unet        & 78.48\%  \\
Deeplabv3p$*$~\cite{chen2018encoder}     & 18'ECCV    &$\mathcal{S}$          & v3p+r101    & 83.63\%  \\
ScribFormer~\cite{li2024scribformer}      & 24'TMI    &$\mathcal{S}$          & ScribFormer & 80.19\%  \\
URSS~\cite{xu2021scribble}             & 21'ICCV   &$\mathcal{S}$          & v2+r101     & 85.23\%  \\
TEL~\cite{liang2022tree}              & 22'CVPR   &$\mathcal{S}$          & v3p+r101    & 85.77\%  \\
AGMM~\cite{wu2023sparsely}             & 23'CVPR   &$\mathcal{S}$          & v3p+r101    & 71.08\%  \\
Ours             & -         &$\mathcal{S}$          & v3p+r101    & \textbf{87.33}\%  \\

\bottomrule
\multicolumn{5}{l}{\footnotesize$^*$: Trained on only scribble annotations with partical cross-entropy loss.}
\end{tabular}
\label{tab:comparison_acdc}
\end{table}
\textbf{Results on ScribbleSup}: In previous conference version, we adopted the pseudo labels generated by BMP~\cite{zhu2023branches}, which was a convolutional-based method trained with only image-level label in stage 1, achieving 75.25\% mIoU with deeplabv2~\cite{chen2017deeplab} and 75.85\% mIoU with DeeplabV3$+$~\cite{chen2018encoder}. In this paper, we improved the original ToCo~\cite{ru2023token}, which was a transformer-based method, with the scribble label in stage 1, outperforming the original ToCo by 1.62\%. It is worth noting that, previous works of ScribbleSup, RAWKS~\cite{vernaza2017learning}, KCL~\cite{tang2018regularized}, and NCL~\cite{tang2018normalized} utilized CRF for postprocessing, which was fairly time-consuming. For recent works of TEL~\cite{liang2022tree} and AGMM~\cite{wu2023sparsely}, they adopted the scribble as dotted line masks for data preprocessing which is unnatural. Therefore, we reimplemented them using standard scribble masks commonly used in previous works~\cite{lin2016scribblesup,tang2018normalized,tang2018regularized,pan2021scribble}. As shown in Table.~\ref{tab:comparison_scribblesup}, our method outperforms all the previous methods, exceeding the TEL by 0.63\% and AGMM by 1.62\%. 
\begin{figure*}
    \includegraphics[width=\linewidth]{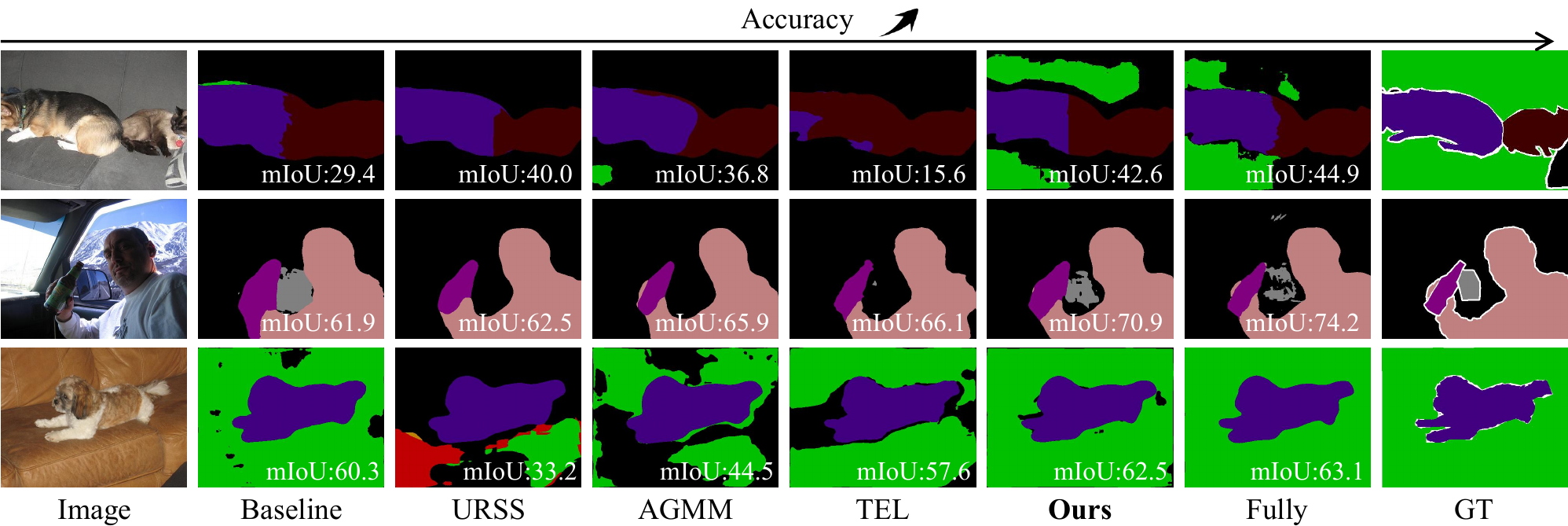}
    \caption{Visual comparison with recent advanced methods on ScribbleSup val set. Our method achieves the best segmentation results. }
    \label{fig:comparevoc-val}
\end{figure*}
\begin{figure*}
    \includegraphics[width=\linewidth]{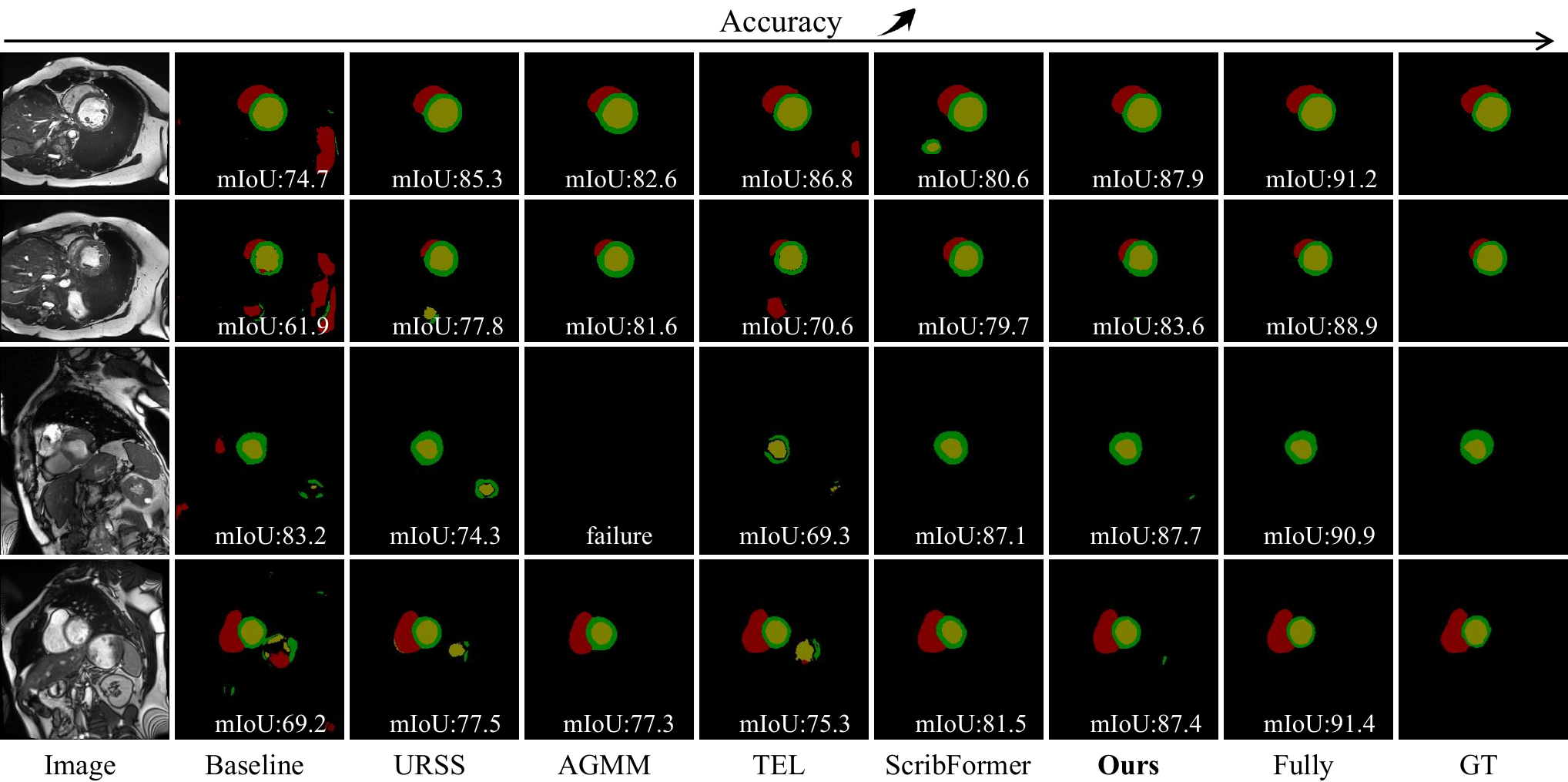}
    \caption{Prediction visualization of different methods on the ScribbleACDC val set. The baseline model is a UNet trained with only the scribble label using cross-entropy loss. Our method is not only effective on natural image datasets, but also performs well on medical image dataset.}
    \label{fig:compareacdc-val}
\end{figure*}

\textbf{Results on ScribbleACDC}: On this dataset, we adopted the resnet~\cite{he2016deep} as the backbone and DeeplabV3$+$ as the segmentor, and the scribble-promoted ToCo is adopted in the first stage to generate pseudo-labels. We also involved the ScribFormer~\cite{li2024scribformer} for comparison, which was proposed specifically for scribble-supervised medical image segmentation. To keep aligned with the medical image segmentation field, we adopted the UNet~\cite{ronneberger2015u} as the baseline model. In this dataset, we train the UNet with only scribble annotations, using the partial cross-entropy loss in Eq.~\ref{eq:l_segs}. ScribFormer introduced the transformer architecture to the UNet, which proved to be more suitable for scribble-based medical image segmentation. We reimplemented these two methods with our split of the dataset. As presented in Table.~\ref{tab:comparison_acdc}, our method exceeds other methods with the mIoU of 87.33\%, showcasing its superiority on medical image segmentation dataset. In comparison, our method is not only applicable to natural image datasets but also competitive on the medical image dataset.

\begin{table*}
\centering
\caption{Ablation study of each componenet on ScribbleCOCO. $\uparrow$ means higher is better, $\downarrow$ means lower is better. }
\begin{tabular}{c|cc|cc|c|ccc|cc} 
\toprule
\multirow{2}{*}{Line number} & \multicolumn{2}{c|}{$\mathcal{L}_{seg}$}         & \multicolumn{2}{c|}{$\mathcal{L}_{dp}$}      & \multirow{2}{*}{$\mathcal{L}_{lorm}$} & \multicolumn{3}{c|}{mIoU}   & \multirow{2}{*}{mean$\uparrow$} & \multirow{2}{*}{std$\downarrow$}  \\ 
\cline{2-5}\cline{7-9}
                             & $\mathcal{L}_{segs}$ & $\mathcal{L}_{segc}$ & $\mathcal{L}_{ds}$ & $\mathcal{L}_{dc}$ &                                            & s1      & s2      & s3      &                                   &                                     \\ 
\hline
1                            & \checkmark                & -                         & -                       & -                       & -                                          & 52.37\% & 52.32\% & 52.18\% & 52.29\%                           & 0.10\%                              \\
2                            & -                         & \checkmark                &                         &                         &                                            & 47.92\% & 48.58\% & 47.80\% & 48.10\%                           & 0.42\%                              \\
3                            & \checkmark                & \checkmark                & -                       & -                       & -                                          & 52.90\% & 53.26\% & 52.94\% & 53.03\%                           & 0.19\%                              \\ 
\hline
4                            & \checkmark                & \checkmark                & \checkmark              & -                       & -                                          & 53.60\% & 54.11\% & 53.78\% & 53.83\%                           & 0.26\%                              \\
5                            & \checkmark                & \checkmark                & -                       & \checkmark              & -                                          & 53.51\% & 53.84\% & 53.59\% & 53.65\%                           & 0.17\%                              \\
6                            & \checkmark                & \checkmark                & \checkmark              & -                       & \checkmark                                 & 53.80\% & 53.95\% & 53.90\% & 53.88\%                           & 0.08\%                              \\
7                            & \checkmark                & \checkmark                & -                       & \checkmark              & \checkmark                                 & 53.74\% & 53.99\% & 53.76\% & 53.83\%                           & 0.13\%                              \\
8                            & \checkmark                & \checkmark                & -                       & -                       & \checkmark                                 & 53.00\% & 53.11\% & 53.00\% & 53.04\%                           & \textbf{0.06}\%                              \\
9                            & \checkmark                & \checkmark                & \checkmark              & \checkmark              & -                                          & 54.11\% & 54.41\% & 54.10\% & 54.21\%                           & 0.18\%                              \\ 
\hline
10                           & \checkmark                & \checkmark                & \checkmark              & \checkmark              & \checkmark                                 & \textbf{54.49}\% & \textbf{54.64}\% & \textbf{54.37}\% & \textbf{54.50}\%                           & 0.14\%                              \\
\bottomrule
\end{tabular}
\label{tab:ablation_scribblecoco}
\end{table*}
\begin{table*}
\centering
\caption{Discussion of pseudo-labels on ScribbleCOCO, mIoU(\%). $\uparrow$ means higher is better, $\downarrow$ means lower is better. }
\begin{tabular}{lp{1.3cm}<{\centering}p{1cm}<{\centering}p{1.5cm}<{\centering}p{1.1cm}<{\centering}p{1.1cm}<{\centering}p{1.1cm}<{\centering}p{1.1cm}<{\centering}p{1.1cm}<{\centering}} 

\toprule
Method  & Pub                   & \textit{Sup} & mIoU$_{pseudo}$ & s1      & s2      & s3      & mean$\uparrow$         & std$\downarrow$        \\ 
\hline
ToCo+s1* & \multirow{3}{*}{-} & $\mathcal{I+S}$ & 46.87\%      & \textbf{54.49}\% & -       & -       & \multirow{3}{*}{\textbf{54.50\%}} & \multirow{3}{*}{0.14\%}  \\
ToCo+s2* &                       & $\mathcal{I+S}$ & 47.60\%      & -       & \textbf{54.64}\% & -       &                          &                          \\
ToCo+s3* &                       & $\mathcal{I+S}$ & 46.22\%      & -       & -       & \textbf{54.37}\% &                          &                          \\ 
\hline
ToCo~\cite{ru2023token}    & CVPR2023              & $\mathcal{I}$   & 42.48\%      & 52.26\% & 52.33\% & 52.32\% & 52.30\%                         & 0.04\%                            \\
ReCAM~\cite{chen2022class}   & CVPR2022              & $\mathcal{I}$   & 45.48\%      & 54.49\% & 54.34\% & 54.30\% & 54.38\%                         & 0.10\%                            \\
CutMix~\cite{yun2019cutmix}  & ICCV2019              & $\mathcal{I}$   & 41.54\%      & 54.06\% & 54.09\% & 53.97\% & 54.04\%                         & \textbf{0.07\%}                            \\
Has~\cite{kumar2017hide}     & ICCV2017              & $\mathcal{I}$   & 41.62\%      & 54.09\% & 54.07\% & 54.01\% & 54.06\%                         & 0.04\%                            \\
\bottomrule

\multicolumn{9}{p{14.5cm}}{*ToCo is improved with the scribble label during the training phase in stage 1. $\mathcal{I}$: supervised by the image-level class label. $\mathcal{S}$: supervised by scribble label. $\mathcal{F}$: fully supervised. Results of s1, s2, and s3 are trained with different styles of scribble label. mIoU$_{pseudo}$: the accuracy of the pseudo-label.}
\end{tabular}
\label{tab:ablation_pseudolabel}
\end{table*}

\subsection{Visual results}
To further demonstrate the effectiveness of our method, we visualized the prediction results of our method along with other milestones on the validation set of each dataset, as presented in Fig.~\ref{fig:comparecoco-train},~\ref{fig:comparecoco-val},~\ref{fig:comparecityscapes-val},~\ref{fig:comparevoc-val},~\ref{fig:compareacdc-val}. Specifically, in Fig.~\ref{fig:comparecoco-train},~\ref{fig:comparecoco-val} and~\ref{fig:comparecityscapes-val}, we presented the predictions on the same sample of each method, where in each row, the scribble annotations for training are randomly generated.

\section{Discussion}
\subsection{The effectiveness of each module}\label{sec:ablation}
The previous conference version~\cite{Scribble023Zhang} has validated the effectiveness of each component with deeplabv2. In this paper, we further validate the components of our method on ScribbleCOCO using DeeplabV3$+$ segmentor with resnet101 backbone, as presented in Table.~\ref{tab:ablation_scribblecoco}. We conducted this experiment 3 times on different scribble styles,  which are denoted as ``s1" (short for scribble 1), ``s2", and ``s3".  From the first three lines in the table, it can be seen that employing either scribble or the pseudo label for basic supervision $\mathcal{L}_{seg}$ will result in an unsatisfied segmentation performance, where utilizing both of them will improve the segmentation performance significantly. Based on this, taking line 3 as the reference, it can be concluded from line 4 to line 9 that, each component can improve the model exclusively, where using any combination of them improves even more. Taking (line 4, line 6) and (line 5, line 7) for comparison, it can be seen that the involvement of the LoRM will significantly improve the stability of the model, generating robust segmentation results on the three scribble styles. Surprisingly in line 8, utilizing the LoRM alone achieves the lowest standard deviation of 0.06\%. This may be attributed to the success of LoRM in rectifying the representations misled by the noisy labels, which equips the backbone with the ability of correctly model the discriminative features for foreground objects. Compared to the baseline, each component obtains a better performance, while using all of them obtains the best performance.

\subsection{The contribution of the pseudo-labels.} 
Unlike previous pseudo-label based methods, where the accuracy of pseudo-labels acts as the upper bound for the model's performance. In our approach, pseudo-labels contribute to the segmentation model by ensuring a baseline level of segmentation accuracy. Rather than merely setting a performance ceiling, these pseudo-labels actively support and reinforce the model's segmentation capabilities. To demonstrate this, we further conducted experiments on ScribbleCOCO with various pseudo-labels to evaluate their impact on model performance, employing DeeplabV3$+$ and the resnet101 backbone as the segmentation framework. We reimplemented the classical image-level weakly supervised semantic segmentation methods to generate the pseudo-labels, including HAS~\cite{kumar2017hide}, Cutmix~\cite{yun2019cutmix}, ReCAM~\cite{chen2022class}, and the original ToCo~\cite{ru2023token}. The segmentation results by merely adopting the pseudo-labels are recorded in the first 4 lines in Table.~\ref{tab:comparison_coco2014}. More results by additionally adopting the scribbles are recorded in Table.~\ref{tab:ablation_pseudolabel}. It has to note that, the ``\textit{Sup}'' stands for the training supervision of the pseudo-label generation methods. Compared with the original ToCo, our strategy of involving the scribble pixel-level supervision has a significant improvement of the pseudo-label, which promoted the original pseudo label mIoU from 42.48\% to 46.22-47.60\% under different scribble styles. The conclusion can be drawn from the tables that, with more accurate pseudo-labels, the accuracy of our method improves significantly, highlighting its compatibility with advancements in image-level weakly supervised semantic segmentation methods.

\subsection{The robustness to the scribble degradation.}
\begin{figure*}
  \centering
  \includegraphics[width=\linewidth]{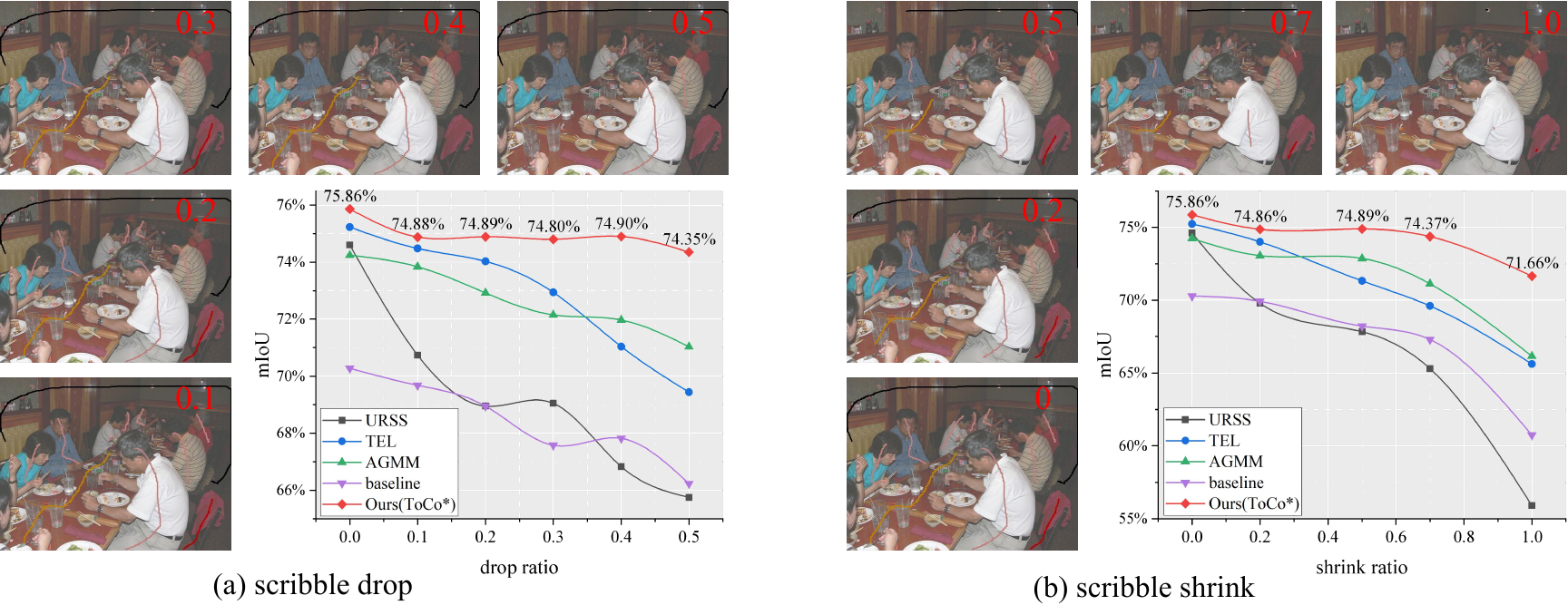}
  \caption{The robustness evaluation of our method on ScribbleSup with different degradation ratios of the scribble. The baseline is a deeplabv3p model with resnet101 backbone, trained on the scribble label exclusively. }
  \label{fig:shrink_drop}
\end{figure*}\begin{figure}
    \centering
    \includegraphics[width=\linewidth]{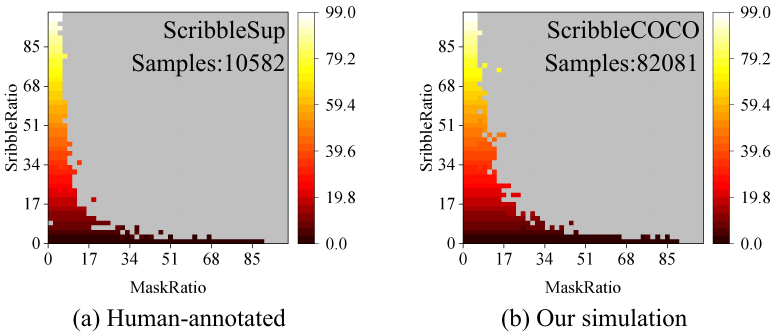}
    \caption{The x-axis represents the proportion of the ground truth masks in the image, the y-axis  represents the proportion of the scribble masks in the ground masks. The units are percentage(\%). These matrix heatmaps reflect the correlations between the mask ratio and the scribble ratio, revealing the habits of human annotators. }
    \label{fig:Heatmaps}
\end{figure}

Given the flexibility of scribble-based annotations, it is common for users to annotate scribbles of varying lengths and occasionally omit certain objects. Therefore, evaluating the model's robustness against different shrink or drop ratios is crucial. Some examples of shrunk or dropped annotations are shown in Fig.~\ref{fig:shrink_drop}. The Scribble-shrink and Scribble-drop annotations are provided by URSS~\cite{pan2021scribble}, which is an extension of the ScribbleSup dataset. The baseline in Fig.~\ref{fig:shrink_drop} is a DeeplabV3$+$ model~\cite{chen2017deeplab} with resent101 backbone trained on the scribble exclusively with partial cross-entropy loss. It can be obviously concluded that the segmentation performance of TEL and AGMM degraded significantly with the improvement of the shrink or drop ratio. Taking the extreme situation as an example, when the shrink ratio increases to 1.0, where the scribble is shrunk into a point, the segmentation performance of TEL and AGMM is decreased by almost 10\%. In comparison, our method decreased by just 4.2\%. Similarly, when the drop ratio is 0.5, leaving half of the objects unlabeled, TEL's performance decreases by approximately 5\%, and AGMM by around 3\%, whereas our method experiences a decrease of just 1.53\%. This robustness can be largely attributed to our design of the LoRM as mentioned in Secton~\ref{sec:ablation}, which rectifies the representation misled by the noisy labels, equipping the model with the ability to correctly model the discriminative features for foreground objects.

\subsection{Scribble simulation consistency with human annotators. }
Scribbles are flexible, requiring careful consideration of annotator preferences and random factors when designing the simulation algorithm. As described in Section~\ref{sec:scribble_simulation}, human annotators tend to trace along object boundaries when dealing with large objects. This behavior can be quantitatively described via the correlation heatmap as presented in Fig.~\ref{fig:Heatmaps}. It can be observed that for small objects, the mask ratio strongly correlates with a higher scribble ratio, as annotators are constrained to draw simple, line-style scribbles. As the mask ratio increases, the randomness in selecting scribble styles (line or boundary) also increases, leading to a weaker correlation with the scribble ratio. Notably, the correlation heatmap of ScribbleCOCO closely mirrors that of ScribbleSup, demonstrating the consistency between our simulation algorithm and human annotator behavior.
\section{Conclusion}
In this paper, we propose a novel method called the class-driven scribble promotion network, designed to address the challenge of scribble-supervised semantic segmentation. To mitigate the issue of noisy labels in pseudo-labeling, we introduce a localization rectification module, which refines feature maps in high-dimensional feature space, along with a distance perception module to identify reliable regions and enhance model certainty. Additionally, we developed a scribble simulation algorithm, creating two new large-scale datasets for scribble-supervised semantic segmentation, named ScribbleCOCO and ScribbleCityscapes. Extensive experiments with various scribble styles demonstrate the superiority of our method compared to existing approaches.


\bibliographystyle{IEEEtran}
\bibliography{ref}
\vspace{-15 mm}
\clearpage
\appendix
\renewcommand\thefigure{\Alph{section}\arabic{figure}}    
\renewcommand\thetable{\Alph{section}\arabic{table}}    
\section{Appendix}
\setcounter{figure}{0}  
\setcounter{table}{0}
\subsection{Math symbol notations.}
The essential notations list is provided in Table.~\ref{tab:math_notations} for quickly searching for the meaning of the notations used in our paper. The \textbf{blod} uppercase characters are used to denote the matrix-valued random variables, and the \textbf{\textit{italic bold}} lowercase characters denote the vectors. The sets are denoted with uppercase greek characters.
\begin{table}[h]
    \centering
    \caption{Math symbol notations for matrix and vector.}
    \begin{tabular}{l|p{5.3cm}} 
    \toprule

    \multicolumn{2}{c}{Notations of matrix}    \\ 
    \hline
    $\mathbf{I}\in\mathbb{R}^{3\times H \times W}$ & The input image matrix. \\
    $\mathbf{F^C}\in\mathbb{R}^{C\times H \times W}$ & Feature embeddings of ToCo. \\
    $\mathbf{F^S}\in\mathbb{R}^{C\times H \times W}$ &  Feature embeddings of DeeplabV3$+$. \\
    $\mathbf{\hat{F}^S} \in \mathbb{R}^{C\times H \times W}$ & The rectified feature map. \\
    $\mathbf{Q}\in\mathbb{R}^{C\times H W}$ & The query matrix. \\
    $\mathbf{K}\in\mathbb{R}^{C\times H W}$ & The key matrix. \\
    $\mathbf{A}\in \mathbb{R}^{HW\times H W}$ & The similarity matrix. \\
    $\mathbf{A'}\in \mathbb{R}^{HW\times H W}$ & The masked similarity matrix. \\
    $\mathbf{M}\in \mathbb{R}^{ H\times W}$ & The foreground 0-1 mask. \\
    $\mathbf{M'}\in \mathbb{R}^{1 \times H W}$ & The flattened foreground mask. \\

    \hline
    \multicolumn{2}{c}{Notations of vector}    \\
    \hline
    $\boldsymbol{k}\in \mathbb{R}^{1\times K}$ & The one-hot class label.     \\
    $\boldsymbol{y}_i\in \mathbb{R}^{1\times K}$ & The one-hot ground truth label.     \\
    $\boldsymbol{\widetilde{y}}_i\in \mathbb{R}^{1\times K}$ & The one-hot pseudo label.  \\ 
    $\boldsymbol{p}_i\in \mathbb{R}^{1\times K}$ & The prediction from the model.  \\ 

    \hline
    \multicolumn{2}{c}{Notations of set}    \\ 
    \hline
    $\mathbf{\Omega}$ & The ground truth label set.    \\
    $\widetilde{\mathbf{\Omega}}$ & The pseudo label set \\
    $\mathbf{\Omega}_{s} \subset \mathbf{\Omega}$ & The scribble annotation label set. \\

    \bottomrule
    \end{tabular}
    \label{tab:math_notations}
\end{table}

\subsection{Details of re-implemented methods}

\begin{figure}[t]
    \subfloat[\centering {Image}]{
        \begin{minipage}[b]{0.15\textwidth}
    	\includegraphics[width=0.8\linewidth]{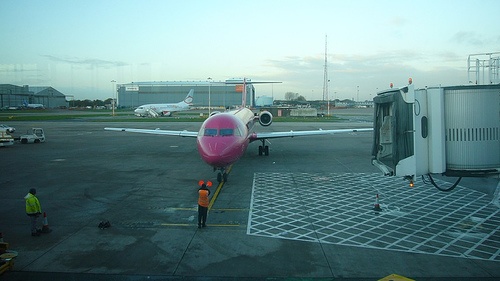}\\
        \includegraphics[width=0.8\linewidth]{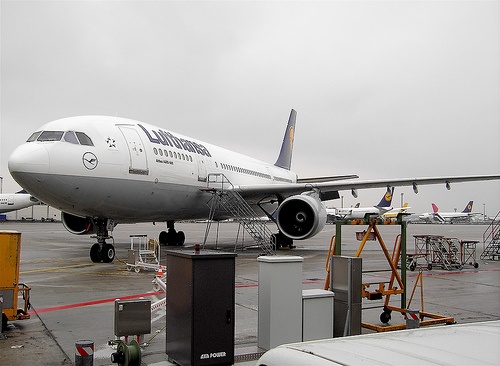}
    	\end{minipage}
    }
    \subfloat[\centering {Common scribble}]{
        \begin{minipage}[b]{0.15\textwidth}
    	\includegraphics[width=0.8\linewidth]{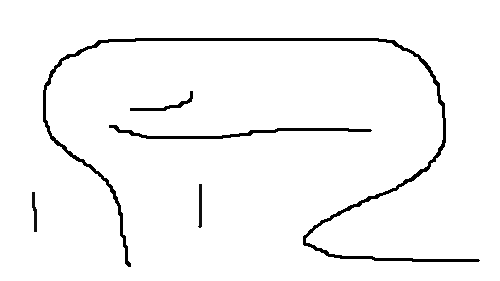} \\
        \includegraphics[width=0.8\linewidth]{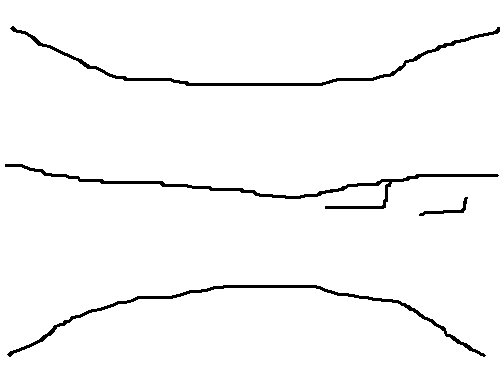}
    	\end{minipage}
    }
    \subfloat[\centering {TEL scribble}]{
        \begin{minipage}[b]{0.15\textwidth}
    	\includegraphics[width=0.8\linewidth]{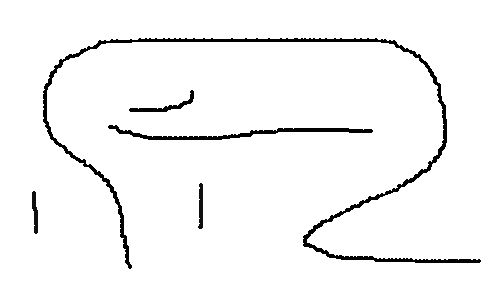}\\
        \includegraphics[width=0.8\linewidth]{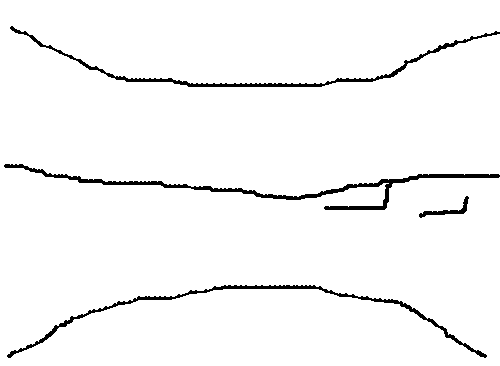}
    	\end{minipage}
    }
    \caption{The scribble annotations comparison used in TEL and AGMM are different than the commonly used ones, which provides extra supervision for training. This may result in unfairness to previous methods. It is recommended to zoom in for better viewing.}
    \label{fig:scribbles}
\end{figure}

\textbf{URSS}~\cite{pan2021scribble} is a scribble-based WSSS method proposed in ICCV 2021, which devises a self-supervised consistency loss to capture the invariant features for advanced segmentation performance. It was trained with sribble annotations that are commonly used in previous scribble-based WSSS~\cite{lin2016scribblesup,tang2018normalized,tang2018regularized} works as shown in Fig.~\ref{fig:scribbles}(e).

\textbf{TEL}~\cite{liang2022tree} is a WSSS method proposed for point-based, scribbles-based and bounding box-based annotations in CVPR2022, which introduced the tree filter technique to weakly-supervised semantic segmentation. It generates scribbles using "xml" files provided by ScribbleSup~\cite{lin2016scribblesup} as shown in Fig.~\ref{fig:scribbles}(c), which draws a circle with a diameter of 3 pixels on each inflection point of the scribble.

\textbf{AGMM}~\cite{wu2023sparsely} is a WSSS method proposed for point-based, scribbles-based and bounding box-based annotations in CVPR2023, which proposes an adaptive gaussian mixture to generate advanced pseudo-labels for training. It utilizes the same scribble annotations as TEL as shown in Fig.~\ref{fig:scribbles}(c).

On average, there are 5053 pixels annotated pixels per image in common scribble annotations. But in TEL scribbles, there are 5563 pixels annotated per image, which exceed the commonly used ones by 510 pixels. Therefore, we re-implement TEL and AGMM with the commonly used scribbles for a fair comparison with previous scribble-based WSSS methods.

\subsection{Table data for shrink and drop }
\begin{figure*}
    \centering
    \subfloat[\centering I+S]{\includegraphics[width=0.13\linewidth]{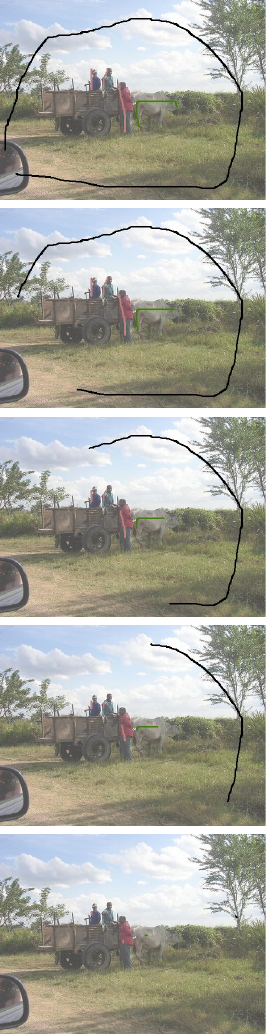}}
    \subfloat[\centering baseline]{\includegraphics[width=0.13\linewidth]{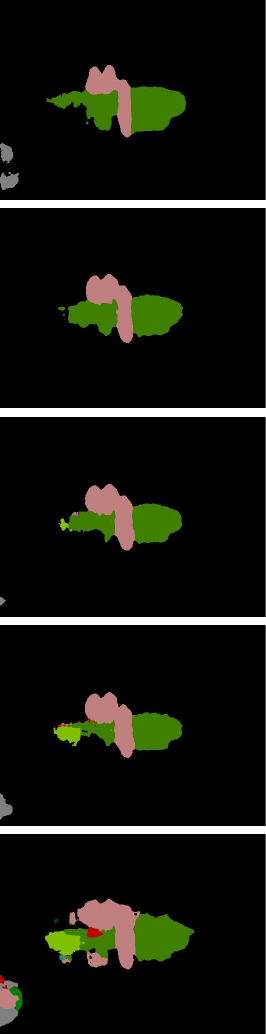}}
    \subfloat[\centering URSS]{\includegraphics[width=0.13\linewidth]{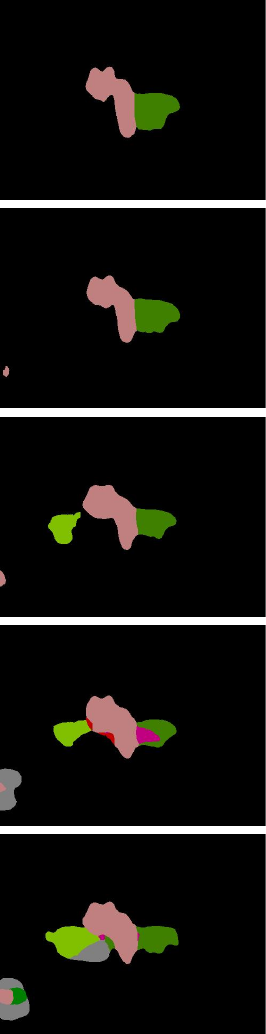}}
    \subfloat[\centering TEL]{\includegraphics[width=0.13\linewidth]{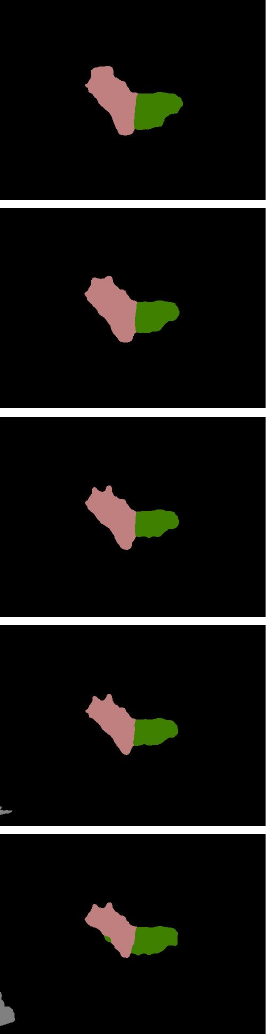}}
    \subfloat[\centering AGMM]{\includegraphics[width=0.13\linewidth]{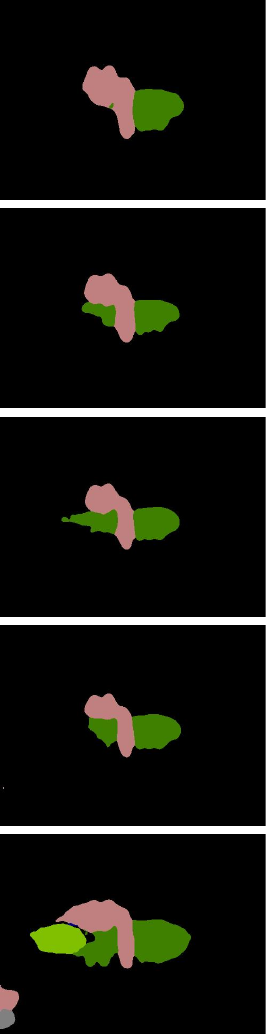}}
    \subfloat[\centering Ours]{\includegraphics[width=0.13\linewidth]{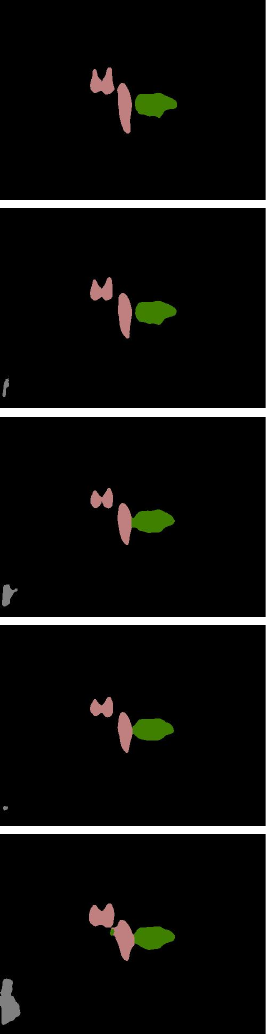}}
    \subfloat[\centering GT]{\includegraphics[width=0.13\linewidth]{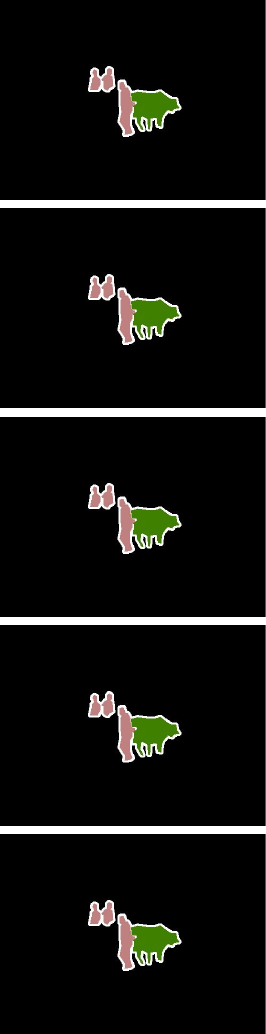}}
    \caption{Visualized comparison on Scribble-shrink. (a) is the image with shrunk scribble annotations, and (g) is the ground truth. From top to down, the shrink ratio is 0, 0.2, 0.5 ,0.7 and 1.}
    \label{figA:viscompare_shrink_1}
\end{figure*}

\begin{figure*}
    \centering
    \subfloat[\centering I+S]{\includegraphics[width=0.13\linewidth]{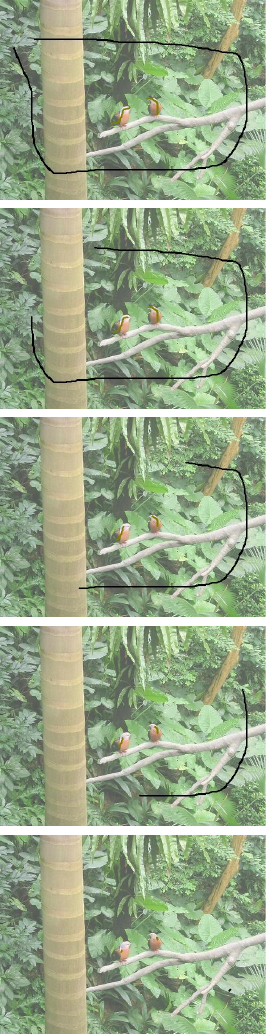}}
    \subfloat[\centering baseline]{\includegraphics[width=0.13\linewidth]{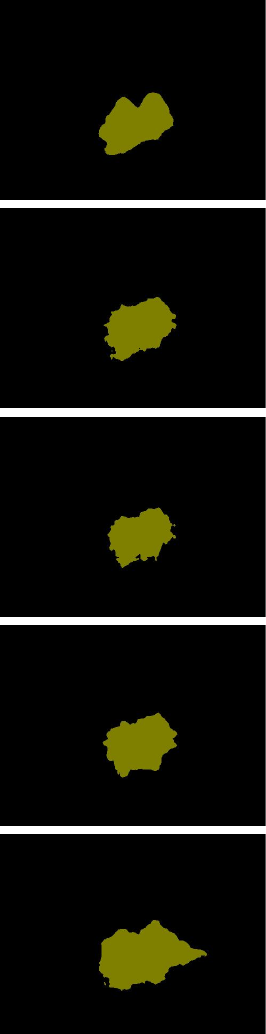}}
    \subfloat[\centering URSS]{\includegraphics[width=0.13\linewidth]{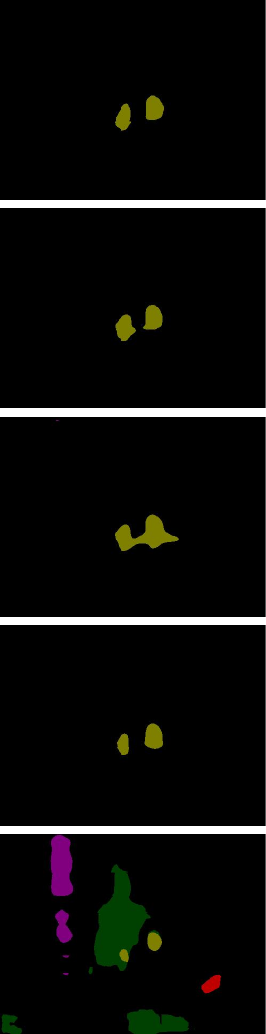}}
    \subfloat[\centering TEL]{\includegraphics[width=0.13\linewidth]{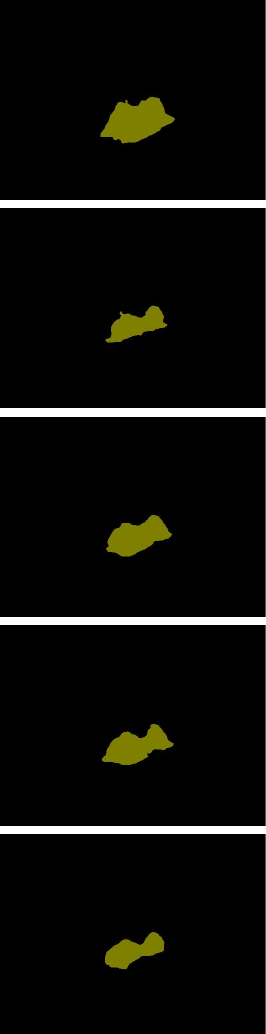}}
    \subfloat[\centering AGMM]{\includegraphics[width=0.13\linewidth]{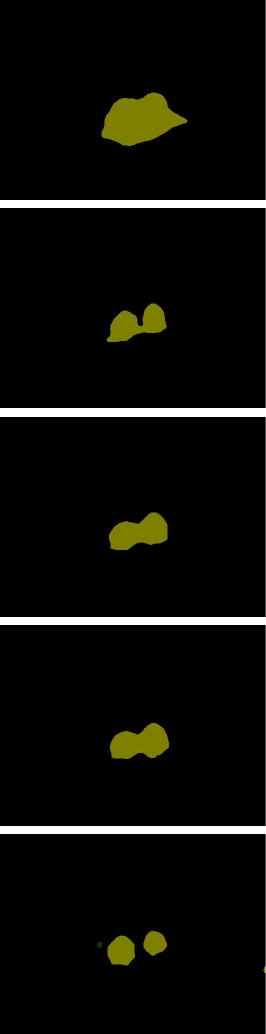}}
    \subfloat[\centering Ours]{\includegraphics[width=0.13\linewidth]{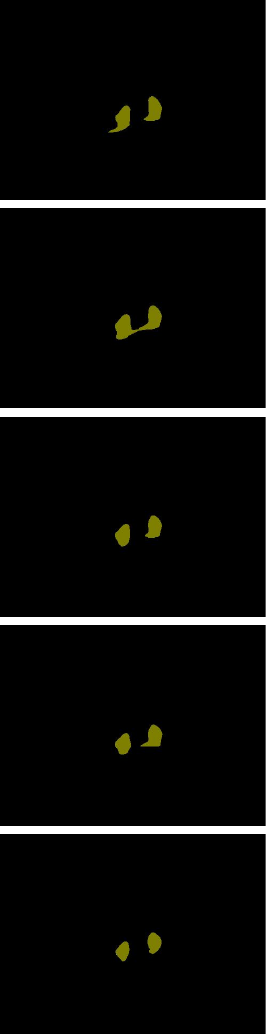}}
    \subfloat[\centering GT]{\includegraphics[width=0.13\linewidth]{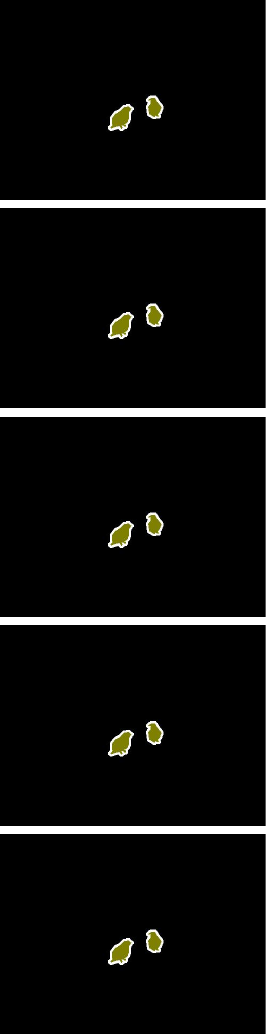}}
    \caption{Visualized comparison on Scribble-shrink. (a) is the image with shrunk scribble annotations, and (g) is the ground truth. From top to down, the shrink ratio is 0, 0.2, 0.5 ,0.7 and 1.}
    \label{figA:viscompare_shrink_2}
\end{figure*}

In Fig.~\ref{fig:shrink_drop} of the main paper, we conduct experiments on \emph{scribble-shrink} and \emph{scribble-drop} datasets provided by URSS~\cite{pan2021scribble}. The detailed qualititatvie results are presented in the following two tables. The baseline model is trained with only scribble annotations employing deeplabV3+ equipped with resnet101 backbone. For the baseline model, which suffers from the problem of overfitting to the sparse annotations, a minor drop or shrink of the annotations may play the role of regularization and improve the performance of the model. More visualization results on \emph{scribble-shrink} and \emph{scribble-drop} are presented in Fig.~\ref{figA:viscompare_shrink_1},~\ref{figA:viscompare_shrink_2},~\ref{figA:viscompare_shrink_1},~\ref{figA:viscompare_shrink_2}.

\begin{table}[h]
\centering
\caption{The data of scribble-drop.}
\begin{tabular}{cccccc} 
\toprule
drop ratio & URSS & TEL     & AGMM    & Ours    & baseline  \\ 
\hline
0        & 74.60\% & 75.23\% & 74.24\% & \textbf{75.86\%} & 70.28\%   \\
0.1~     & 70.74\% & 74.48\% & 73.84\% & \textbf{74.88\%} & 69.68\%   \\
0.2~     & 68.95\% & 74.03\% & 72.92\% & \textbf{74.89\%} & 68.95\%   \\
0.3~     & 69.05\% & 72.94\% & 72.15\% & \textbf{74.80\%} & 67.58\%   \\
0.4~     & 66.83\% & 71.04\% & 71.97\% & \textbf{74.90\%} & 67.82\%   \\
0.5~     & 65.75\% & 69.44\% & 71.03\% & \textbf{74.35\%} & 66.23\%   \\
\bottomrule
\end{tabular}
\end{table}

\begin{table}[h]
\centering
\caption{The data of scribble-shrink in Figure.7(b)}
\begin{tabular}{cccccc} 
\toprule
shrink ratio & URSS & TEL     & AGMM    & Ours    & baseline  \\ 
\hline
0          & 74.60\% & 75.23\% & 74.24\% & \textbf{75.86\%} & 70.28\%   \\
0.2        & 69.78\% & 74.00\% & 73.06\% & \textbf{74.86\%} & 69.92\%   \\
0.5        & 67.84\% & 71.32\% & 72.86\% & \textbf{74.89\%} & 68.22\%   \\
0.7        & 65.30\% & 69.61\% & 71.12\% & \textbf{74.37\%} & 67.30\%   \\
1          & 55.90\%  & 65.62\% & 66.15\% & \textbf{71.66\%} & 60.74\%   \\
\bottomrule
\end{tabular}
\end{table}


\begin{figure*}
    \centering
    \subfloat[\centering I+S]{\includegraphics[width=0.13\linewidth]{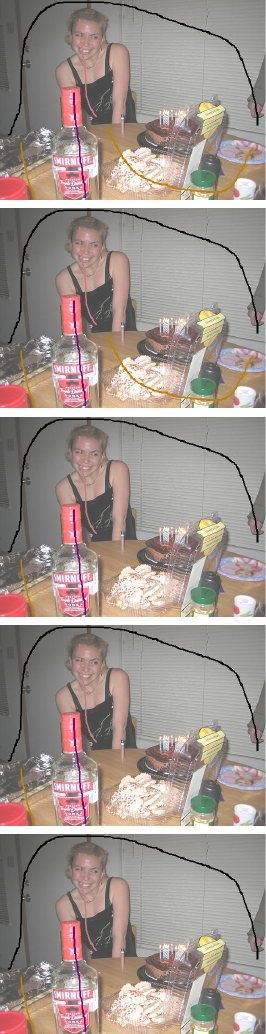}}
    \subfloat[\centering baseline]{\includegraphics[width=0.13\linewidth]{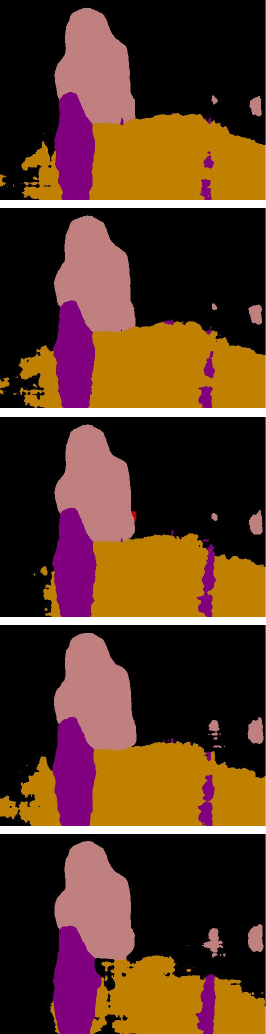}}
    \subfloat[\centering URSS]{\includegraphics[width=0.13\linewidth]{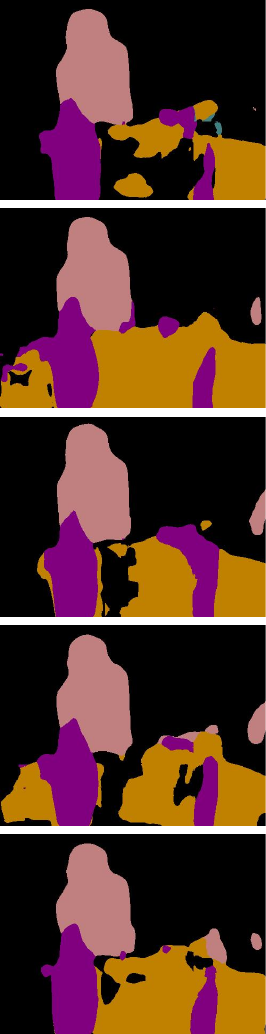}}
    \subfloat[\centering TEL]{\includegraphics[width=0.13\linewidth]{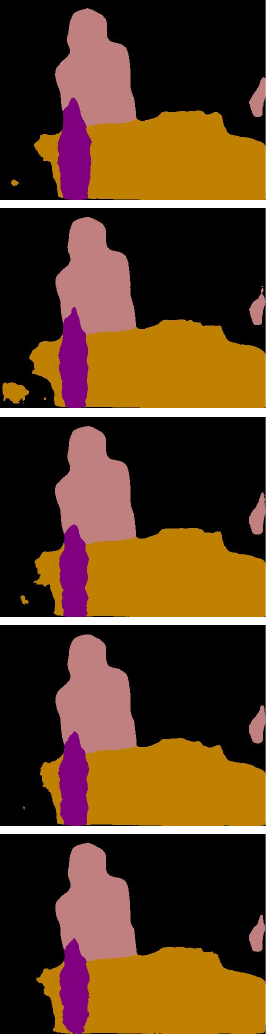}}
    \subfloat[\centering AGMM]{\includegraphics[width=0.13\linewidth]{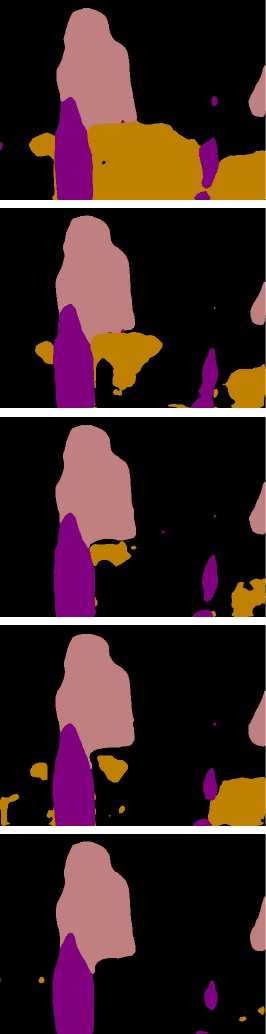}}
    \subfloat[\centering Ours]{\includegraphics[width=0.13\linewidth]{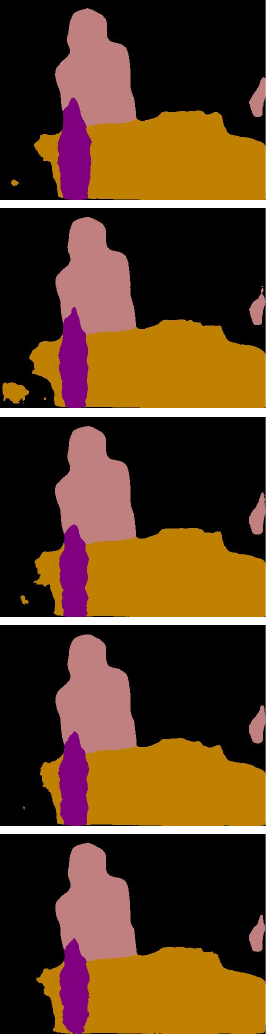}}
    \subfloat[\centering GT]{\includegraphics[width=0.13\linewidth]{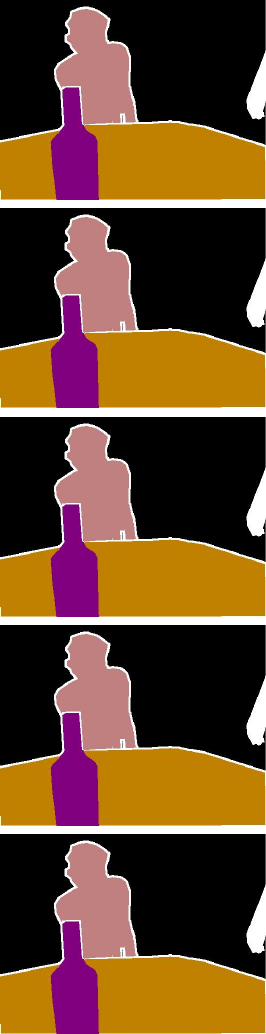}}
    \caption{Visualized comparison on Scribble-drop. (a) is the image with dropped scribble annotations, and (g) is the ground truth. From top to down, the drop ratio is 0.1, 0.2, 0.3 ,0.4, and 0.5.}
    \label{figA:viscompare_drop_1}
\end{figure*}
\begin{figure*}
    \centering
    \subfloat[\centering I+S]{\includegraphics[width=0.13\linewidth]{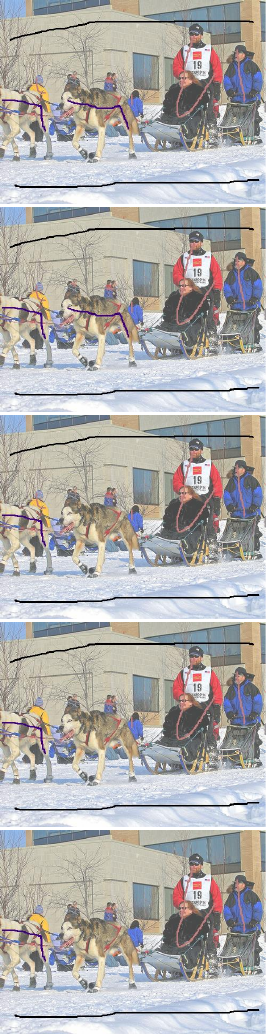}}
    \subfloat[\centering baseline]{\includegraphics[width=0.13\linewidth]{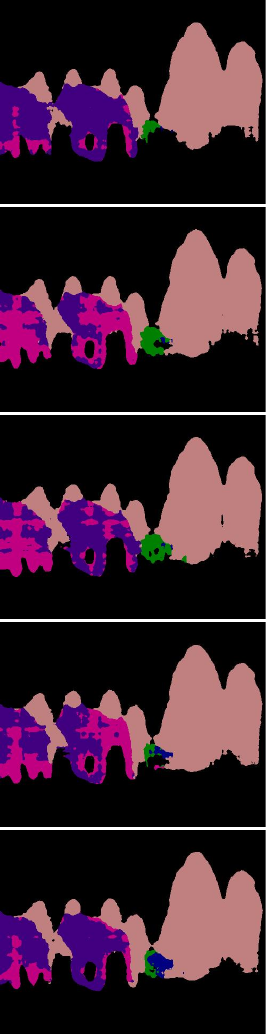}}
    \subfloat[\centering URSS]{\includegraphics[width=0.13\linewidth]{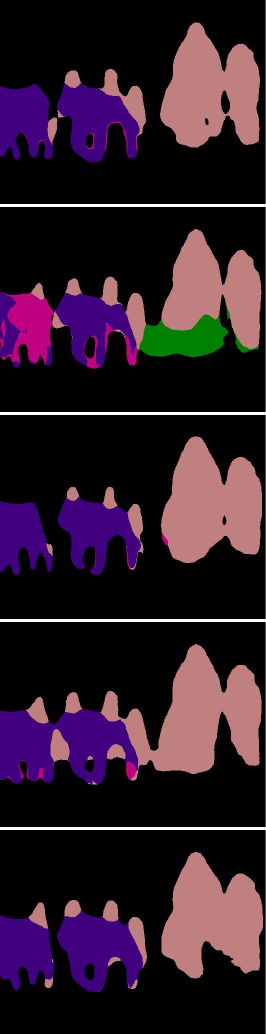}}
    \subfloat[\centering TEL]{\includegraphics[width=0.13\linewidth]{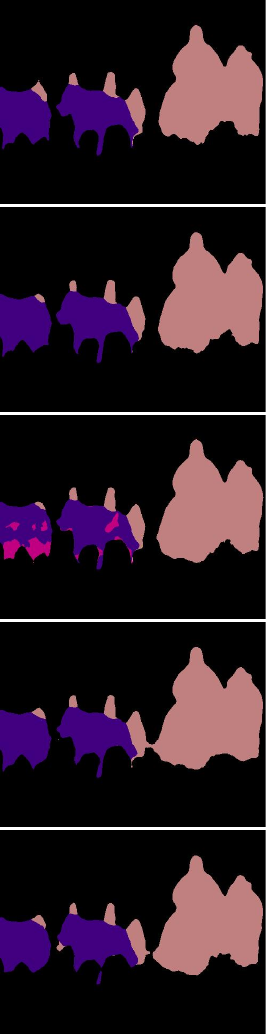}}
    \subfloat[\centering AGMM]{\includegraphics[width=0.13\linewidth]{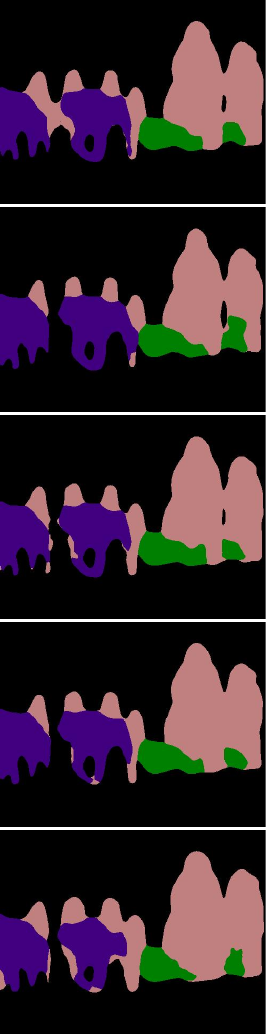}}
    \subfloat[\centering Ours]{\includegraphics[width=0.13\linewidth]{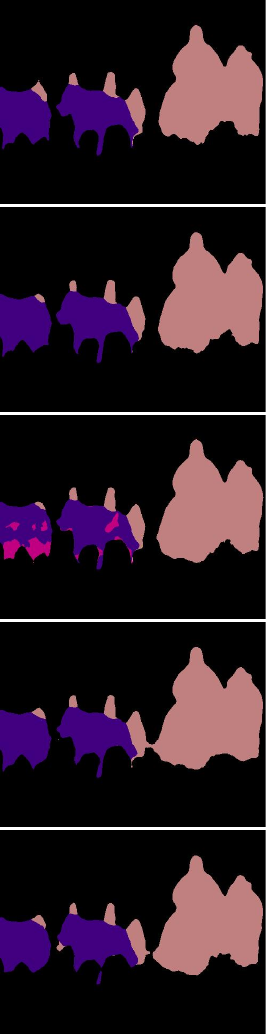}}
    \subfloat[\centering GT]{\includegraphics[width=0.13\linewidth]{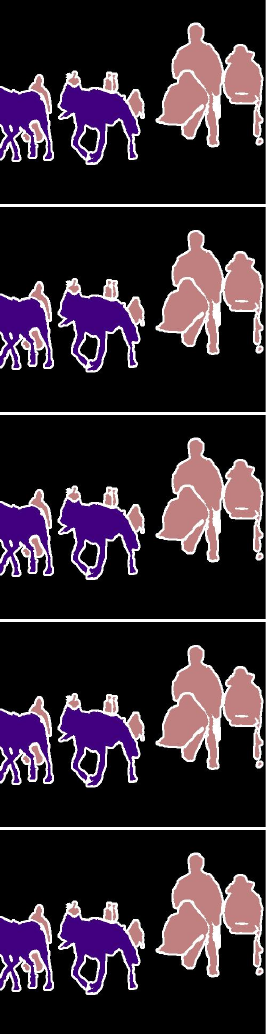}}
    \caption{Visualized comparison on Scribble-drop. (a) is the image with dropped scribble annotations, and (g) is the ground truth. From top to down, the drop ratio is 0.1, 0.2, 0.3 ,0.4, and 0.5.}
    \label{figA:viscompare_drop_2}
\end{figure*}
\clearpage
\subsection{Hyper-parameters for training}

In stage 1, we utilized the Distributed Data Parallel (DDP) function in PyTorch to train the scribble-promoted ToCo. For different datasets, the training hyper-parameters are listed in Table.~\ref{tab:toco_parameter}. 

In stage 2, the DeeplabV3$+$~\cite{chen2018encoder} was adopted as the segmentor with resnet101~\cite{he2016deep} as the backbone. The training hyper-parameters for each dataset were listed in Table.~\ref{tab:cdsp_params}.

\begin{table}
\centering
\caption{Hyper-parameters for training scribble-promoted ToCo}
\begin{tabular}{lccccc} 
\toprule
Dataset            & \multicolumn{1}{l}{GPUs} & \multicolumn{1}{l}{SPG$^*$} & \multicolumn{1}{l}{Iters} & \multicolumn{1}{l}{$\lambda_{cls}$} & \multicolumn{1}{l}{$\lambda_{segs}$}  \\
\hline
ScribbleSup        & 2                        & 2                       & 20000                     & 1                             & 0.03                           \\
ScribbleCOCO       & 4                        & 2                       & 80000                     & 1                             & 0.05                           \\
ScribbleCityscapes & 2                        & 4                       & 20000                     & 1                             & 0.5                            \\
ScribbleACDC               & 2                        & 2                       & 9000                      & 0.1                           & 0.9                            \\
\bottomrule
\multicolumn{6}{l}{* \footnotesize SPG: short for samples per GPU}
\end{tabular}
\label{tab:toco_parameter}
\end{table}
\begin{table}
\centering
\caption{Hyper-parameters for training deeplabv3+.}
\begin{tabular}{lccccc} 
\toprule
Dataset            & learning rate   & \multicolumn{1}{l}{$\lambda_{segc}$} & \multicolumn{1}{l}{$\lambda_{lorm}$} & \multicolumn{1}{l}{$\lambda_s$} & \multicolumn{1}{l}{$\lambda_c$}  \\ 
\hline
ScribbleSup        & 2e-3 & 0.8                            & 0.5                                   & 1                              & 7                               \\
ScribbleCOCO       & 2e-3 & 0.8                            & 0.5                                   & 1                              & 7                               \\
ScribbleCityscapes & 1e-2 & 0.4                            & 0.5                                   & 2                              & 7                               \\
ScribbleACDC               & 5e-2 & 0.1                            & 0.5                                   & 1                              & 6                               \\
\bottomrule
\end{tabular}
\label{tab:cdsp_params}
\end{table}
\subsection{Visualization of ScribbleCOCO and ScribbleCityscapes}
\begin{figure*}[p]
    \centering
    \includegraphics[width=0.9\linewidth]{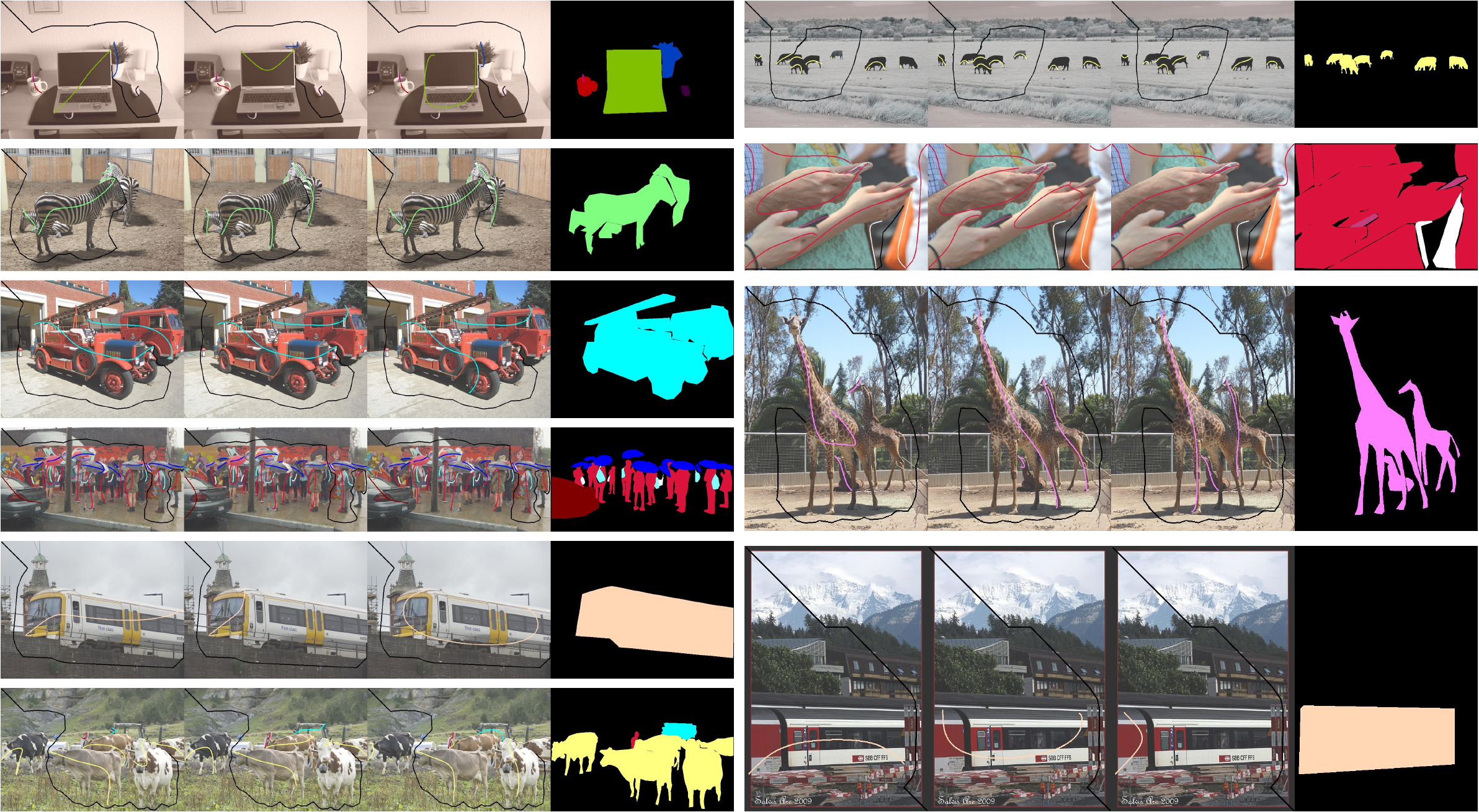}
    
    \includegraphics[width=0.9\linewidth]{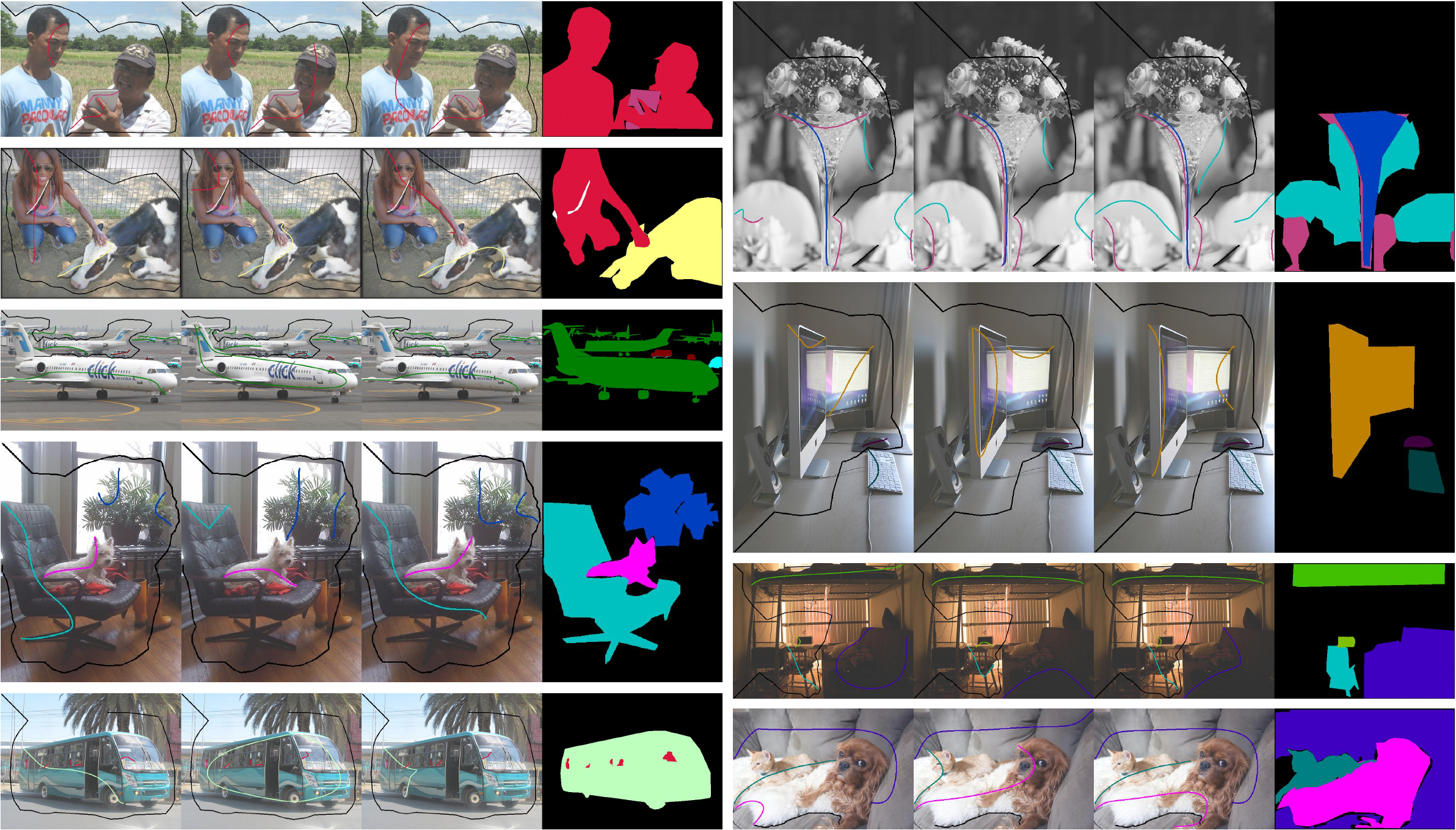}
    \caption{Visualized examples from ScribbleCOCO. For each group, the left three are randomly generated scribbles, the right one is the ground truth mask. }
    \label{figA:visScribbleCOCO}
\end{figure*}

\begin{figure*}
    \centering
    \includegraphics[width=0.9\linewidth]{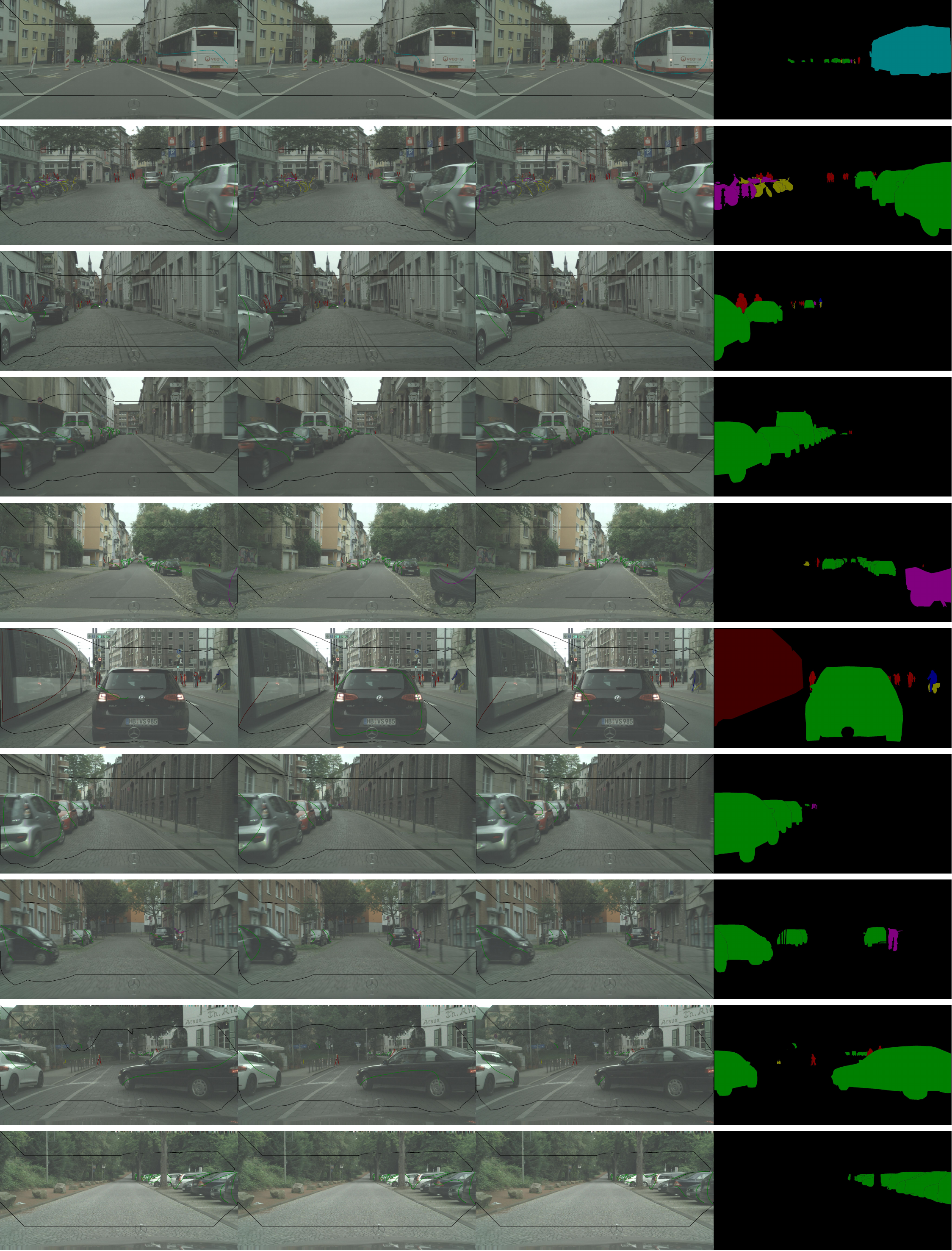}
    \caption{Visualized examples from ScribbleCitycapes. For each group, the left three are randomly generated scribbles, the right one is the ground truth mask. }
    \label{figA:visScribbleCityscapes}
\end{figure*}

ScribbleCOCO and ScribbleCityscapes are new datasets proposed by this paper. On these datasets, we generated the scribble annotations 3 times with our scribble simulation algorithm to acquire random scribble styles. Visualization examples are presented in Fig.~\ref{figA:visScribbleCOCO} and Fig.~\ref{figA:visScribbleCityscapes}.

\end{document}